%% file: main.tex
\documentclass[10pt,journal,compsoc]{IEEEtran}

\input{packages}

\input{definitions}

\begin{document}

\title{Few-Shot Electronic Health Record Coding through Graph Contrastive Learning}

\input{author}

\markboth{Journal of the IEEE Transactions on Knowledge and Data Engineering,~Vol.~x, No.~x, July~2021}%
{Shell \MakeLowercase{\textit{et al.}}: Bare Advanced Demo of IEEEtran.cls for IEEE Computer Society Journals}

\IEEEtitleabstractindextext{%
\input{00-abstract}
\begin{IEEEkeywords}
Electronic health record (EHR) coding, few-shot learning, graph contrastive learning.
\end{IEEEkeywords}}

\maketitle

\IEEEdisplaynontitleabstractindextext
\IEEEpeerreviewmaketitle

\input{01-introduction}
\input{02-relatedWork}
\input{03-preliminaries}

\input{04-method}
\input{05-evaluation}
\input{06-results}
\input{07-analysis}

\input{08-conclusion}
\input{acknowledge}

\ifCLASSOPTIONcaptionsoff
  \newpage
\fi

\bibliographystyle{plainnat}
\bibliography{main.bbl}

\end{document}

%% file: packages.tex
\usepackage[numbers]{natbib}
\usepackage{amsmath}
\usepackage{amssymb}
\usepackage{acronym}
\usepackage{hyperref}
\usepackage{booktabs} 
\usepackage[skip=0pt]{caption}
\usepackage{multirow}
\usepackage{bm}
\usepackage{bbm}
\usepackage{balance}
\usepackage{lineno}
\usepackage{subfig}
\usepackage[inline]{enumitem}
\usepackage{pifont}
\usepackage{array}
\usepackage{graphicx}
\usepackage[ruled,commentsnumbered,linesnumbered]{algorithm2e}
\usepackage[T1]{fontenc}
\usepackage{aecompl}
\usepackage{color}

%% file: definitions.tex
\AtBeginDocument{%
  \providecommand\BibTeX{{%
    \normalfont B\kern-0.5em{\scshape i\kern-0.25em b}\kern-0.8em\TeX}}}
\makeatletter
\renewcommand\normalsize{%
\@setfontsize\normalsize\@xpt\@xiipt
\abovedisplayskip 10\p@ \@plus2\p@ \@minus5\p@
\abovedisplayshortskip \z@ \@plus3\p@
\belowdisplayshortskip 10\p@ \@plus3\p@ \@minus3\p@
\belowdisplayskip \abovedisplayskip
\let\@listi\@listI}
\makeatother

\acrodef{EHR}{electronic health record}
\acrodef{ICD}{International Classification of Diseases}
\acrodef{FSL}{few-shot learning}
\acrodef{HEWE}{heterogeneous \ac{EHR} word-entity}
\acrodef{GSCL}{graph sampling contrastive learning}
\acrodef{GECL}{graph evolution contrastive learning}
\acrodef{GCC}{graph contrastive coding}
\acrodef{TF-IDF}{term frequency-inverse document frequency}
\acrodef{PMI}{point-wise mutual information}
\acrodef{GCN}{graph convolutional network}
\acrodef{MLP}{multilayer perceptron}
\acrodef{GRU}{gated recurrent unit}
\acrodef{BI-GRU}{bidirectional gate recurrent unit}
\acrodef{AUC}{area under the curve}

\newcommand{\OurMethod}{CoGraph}

%% file: author.tex
\author{Shanshan~Wang,
        Pengjie~Ren,
        Zhumin~Chen$^*$,
        Zhaochun~Ren,
        Huasheng~Liang,
        Qiang~Yan,
        Evangelos~Kanoulas,
        and~Maarten de~Rijke
\thanks{$^*$ Corresponding author.}
\IEEEcompsocitemizethanks{
\IEEEcompsocthanksitem Shanshan Wang, Pengjie Ren, Zhumin Chen and Zhaochun Ren are with Shandong University, Qingdao, China.\\
E-mails: wangshanshan5678@gmail.com, jay.ren@outlook.com, chenzhumin@sdu.edu.cn and zhaochun.ren@sdu.edu.cn.
\IEEEcompsocthanksitem Huasheng Liang and Qiang Yan are with  WeChat, Tencent,
Guangzhou, China.\\
E-mails: watsonliang@tencent.com and rolanyan@tencent.com.
\IEEEcompsocthanksitem
Evangelos Kanoulas and Maarten de Rijke are with University of Amsterdam, Amsterdam, The Netherlands.\\
E-mails: e.kanoulas@uva.nl and m.derijke@uva.nl.

}

}








%% file: 00-abstract.tex

\begin{abstract}
\Acf{EHR} coding is the task of assigning \ac{ICD} codes to each \ac{EHR}. 
Most previous studies either only focus on the frequent \ac{ICD} codes or treat rare and frequent \ac{ICD} codes in the same way.
These methods perform well on frequent \ac{ICD} codes but due to the extremely unbalanced distribution of \ac{ICD} codes, the performance on rare ones is far from satisfactory.
We seek to improve the performance for both frequent and rare \ac{ICD} codes by using a contrastive graph-based \ac{EHR} coding framework, \OurMethod{}, which re-casts \ac{EHR} coding as a few-shot learning task.
First, we construct a \acfi{HEWE} graph for each \ac{EHR}, where the words and entities extracted from an \ac{EHR} serve as nodes and the relations between them serve as edges.
Then, \OurMethod{} learns similarities and dissimilarities between \ac{HEWE} graphs from different \ac{ICD} codes so that information can be transferred among them.
In a few-shot learning scenario, the model only has access to frequent \ac{ICD} codes during training, which might force it to encode features that are useful for frequent \ac{ICD} codes only.
To mitigate this risk, \OurMethod{} devise two graph contrastive learning schemes, \ac{GSCL} and \ac{GECL}, that exploit the \ac{HEWE} graph structures so as to encode transferable features.
\ac{GSCL} utilizes the intra-correlation of different sub-graphs sampled from \ac{HEWE} graphs while \ac{GECL} exploits the inter-correlation among \ac{HEWE} graphs at different clinical stages.

Experiments on the MIMIC-III benchmark dataset show that \OurMethod{} significantly outperforms state-of-the-art methods on \ac{EHR} coding, not only on frequent \ac{ICD} codes, but also on rare codes, in terms of several evaluation indicators.
On frequent \ac{ICD} codes, \ac{GSCL} and \ac{GECL} improve the classification accuracy and F1 by 1.31\% and 0.61\%,  respectively, and on rare \ac{ICD} codes \OurMethod{} has  more obvious improvements by 2.12\% and 2.95\%.
\end{abstract}

%% file: 01-introduction.tex

\section{Introduction}
\IEEEPARstart{T}{he} \acfi{ICD}\footnote{\url{https://www.cdc.gov/nchs/icd/icd9cm.htm}} is a globally used diagnostic tool for epidemiology, health management and clinical purposes that is commonly used in clinical practice to facilitate billing activities, epidemiology assessment and so on~\citep{mlhc/ChoiBSSS16,DBLP:conf/bibm/AvatiJHDNS17}.
Automatic \acfi{EHR} coding is the process of assigning \ac{ICD} codes to \acp{EHR}.
The task is increasingly attracting attention as manual \ac{EHR} coding is time-consuming and error-prone.

Automatic \ac{EHR} coding is non-trivial because the {ICD} code space is large and their distribution is extremely imbalanced.
For instance, 4,065 out of 6,879 \ac{ICD} codes appear less than 10 times in the MIMIC-III dataset~\citep{johnson2016mimic}.
Frequent \ac{ICD} codes are generally easier to predict using machine-learning based methods than less frequent ones.
Due to a lack of training samples, the performance of predicting rare \ac{ICD} codes is far from satisfactory.
Most automated \ac{EHR} coding methods focus on the top 50 or 100 most frequent \ac{ICD} codes and ignore less frequent \ac{ICD} codes \cite[e.g.,][]{DBLP:conf/mlhc/XuLPGBMPKCMXX19,DBLP:journals/cmpb/HuangOS19,DBLP:journals/tkde/WangCLLYS16}. 
Such methods are not very practical as the rare \ac{ICD} codes form a large fraction of the total number. 
In contrast, in some studies~\citep[e.g.,][]{DBLP:conf/naacl/MullenbachWDSE18,DBLP:conf/aaai/BaumelNCEE18}, rare \ac{ICD} codes are treated in the same way as frequent ones.
However, such methods perform well only for frequent \ac{ICD} codes.
A few studies aim to improve the prediction performance on rare \ac{ICD} codes through data augmentation~\citep[e.g.,][]{DBLP:conf/bionlp/ZhangHZL17,frontiers/Teng2020}.
They enrich the information of rare \ac{ICD} codes by introducing external knowledge or synthesizing data. 
Hence, they rely heavily on the quality of external knowledge sources or synthetic data. 
However, such knowledge or synthetic data is generally difficult to obtain.

We propose an alternative, graph based few-shot \ac{EHR} coding approach, called \OurMethod{} (\textbf{Co}ntrastive \textbf{Graph} Coding), to improve the \ac{EHR} coding performance for both \emph{frequent} and \emph{rare} \ac{ICD} codes.
First, we construct a \acfi{HEWE} graph for each \ac{EHR}, where words and entities extracted from an \ac{EHR} serve as nodes.
The relations (i.e., containing relation between \ac{EHR} and words, co-occurrence relation of words, mention relation of entity in Wiki\-pe\-dia) serve as edges. 
Then, for each individual \ac{HEWE} graph, the \ac{ICD} prediction task for each \ac{ICD} code is treated as a meta-task where the probability represents the likelihood that the \ac{ICD} will be assigned to one \ac{HEWE} graph.
By switching between different meta-tasks constructed from frequent \ac{ICD} codes during training, \OurMethod{} allows us to extract transferable information across different \ac{ICD} codes.
As a result, \OurMethod{} cannot only improve the prediction of frequent \ac{ICD} codes;
it can also boost the performance of rare \ac{ICD} codes with just a small number of labeled \ac{HEWE} graphs, by using prior information transferred from non-overlapping frequent \acp{ICD} that have sufficient \ac{HEWE} graphs.

This is not enough. In a few-shot classification scenario, learning methods only have access to instances belonging to frequent classes. 
Without correction this may push a learning method to only encode features that are useful for distinguishing frequent \acp{ICD}, while discarding those that might be critical for rare \acp{ICD}~\citep{sbai2020impact,DBLP:journals/corr/abs-1910-03560}.
To mitigate this risk, \OurMethod{} leverages self-supervised representation learning techniques \cite{DBLP:conf/cvpr/YeZYC19,DBLP:conf/icml/ChenK0H20} for \ac{HEWE} graphs by limiting the dependence of the framework on the \ac{ICD} codes. 
To this end, we propose two graph contrastive learning schemes to exploit information available from the \ac{HEWE} graph structure itself to learn better transferable features, namely \acfi{GSCL} and \acfi{GECL}.

\begin{figure}[t]
\centering  
\includegraphics[width=1\columnwidth]{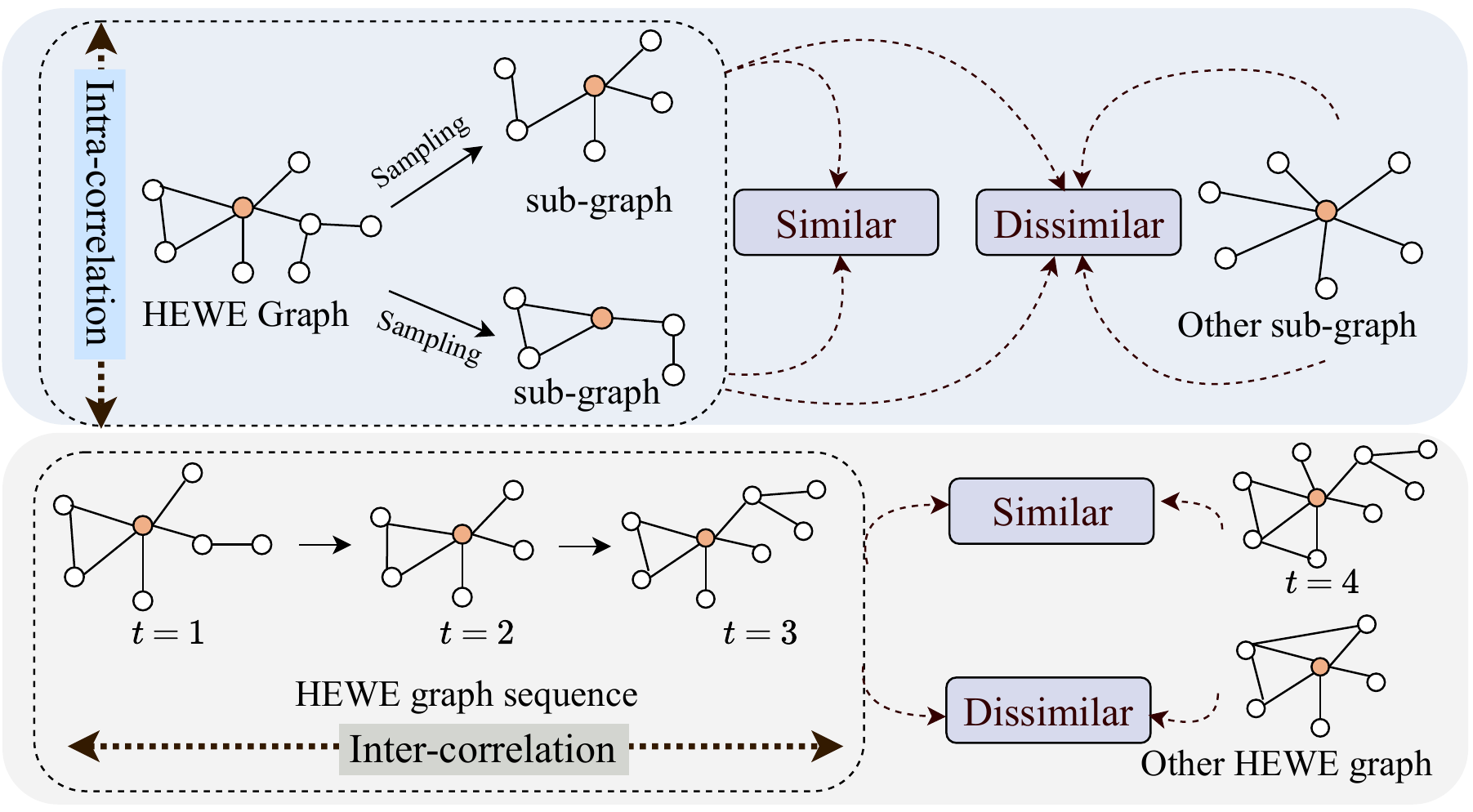}  
\caption{(Top): A diagram of the contrast between the sampled sub-graphs from initial \ac{HEWE} graphs. (Bottom): A diagram of the contrast between a historical \ac{HEWE} graph sequence and future graph at different clinical stages. The orange node is a virtual \ac{EHR} node. $t$ is the $t$-th \ac{HEWE} graph of the same patient.}
\label{intro}
\end{figure}

\ac{GSCL} explores the \emph{intra}-correlation of sub-graphs sampled from the \ac{HEWE} graphs, as illustrated in Fig.~\ref{intro} (top).
The intuition behind this is that each sampled sub-graph is a part of the whole \ac{HEWE} graph, hence the sub-graphs are contextually relevant to each other if they come from the same {HEWE} graph.
The representation of the sub-graph is thus encouraged to capture the shared structural information inherited from the \ac{HEWE} graph.
In contrast, \ac{GECL} exploits the \emph{inter}-correlation among \ac{HEWE} graphs at different clinical stages, as illustrated in Fig.~\ref{intro} (bottom).
The intuition behind it is that \ac{HEWE} graphs from the same patient can be expressed as an evolving graph sequence and display temporal coherence and consistency. 
The history such a \ac{HEWE} graph sequence together with the future \ac{HEWE} graph are considered as similar pairs while the history with another \ac{HEWE} graph picked from other sequences are considered to be dissimilar ones. 
Hence, by learning invariant, slowly changing features in an evolutionary \ac{HEWE} graph sequence, a model will learn to reserve more important features in the \ac{HEWE} graphs. 

Because the two graph contrastive learning schemes, \ac{GSCL} and \ac{GECL}, do not rely on the \ac{ICD} labels for each \ac{EHR} as they define positive and negative samples by contrast, they can help encode unbiased transferable features.
Moreover, by extending the size and variety of training data in this manner, we expect \OurMethod{} to learn stronger \ac{HEWE} graph features and achieve performance gains.
To verify the effectiveness of \OurMethod{} and compare \ac{GSCL} and \ac{GECL} schemes, we conduct experiments on the MIMIC-III benchmark dataset~\citep{johnson2016mimic}.
The experimental results demonstrate that \OurMethod{} significantly outperforms state-of-the-art methods by a large margin, and that two graph contrastive learning schemes can further improve model's performance.

To sum up, the contributions of this work are as follows:
\begin{itemize}[leftmargin=*,nosep]
\item We introduce \OurMethod{}, a graph-based few-shot \ac{EHR} coding framework for boosting the prediction capability on both frequent and rare \ac{ICD} codes. 

\item We devise \emph{\acl{GSCL}} and \emph{\acl{GECL}} schemes by exploring the intra-corre\-lation of each \ac{HEWE} graph and the inter-correlation among sequential \ac{HEWE} graphs at different clinical stages.

\item We carry out extensive experiments on the MIMIC-III benchmark dataset to verify and analyze the effectiveness of \OurMethod{}.
\end{itemize}

%% file: 02-relatedWork.tex

\section{Related Work}
Related work comes in three directions:
\begin{enumerate*}
\item automatic \ac{EHR} coding,
\item few-shot learning, and
\item graph contrastive learning.
\end{enumerate*}

\subsection{Automatic \ac{EHR} coding}
In recent years, the automatic \ac{EHR} coding task has been studied extensively. 
Most \ac{EHR} coding methods focus on (some) frequent \ac{ICD} codes and ignore rare \ac{ICD} codes.
\citet{DBLP:journals/cmpb/HuangOS19} perform two data preprocessing methods; the first only considers the top 10 and top 50 common \ac{ICD} codes; the second groups \ac{ICD} codes into categories based on their hierarchical nature; they extract features from \acp{EHR} and train an \ac{ICD} prediction model using deep learning.
\citet{DBLP:journals/tkde/WangCLLYS16} use the hierarchical nature of \ac{ICD} codes and divide them into 19 categories at the upper level for the most general classification and 129 categories at the lower level for more specific classifications. 
\citet{DBLP:conf/mlhc/XuLPGBMPKCMXX19} develop an ensemble-based approach that integrates three modality-specific models to predict 32 selected \ac{ICD} codes. 
\citet{DBLP:journals/corr/abs-1711-04075} propose a soft-attention mechanism that learns to allocate attention strengths on multiple diagnosis descriptions when assigning the 50 most frequent \ac{ICD} codes.

As we pointed out in the introduction, the frequency distribution of \ac{ICD} codes is imbalanced; rare \ac{ICD} codes account for a large proportion. 
Some \ac{EHR} coding methods treat rare \ac{ICD} and frequent \ac{ICD} codes equally.
\citet{DBLP:conf/naacl/MullenbachWDSE18} aggregate information across a \ac{EHR} by using a convolutional neural network and use an attention mechanism to select the most relevant segments for each  \ac{ICD} code.
Similarly,~\citet{DBLP:conf/aaai/BaumelNCEE18} present a Hierarchical Attention-bidirectional Gated Recurrent Unit (HA-GRU), a hierarchical approach to tag an \ac{EHR} by identifying the sentences relevant for each \ac{ICD} code.
\citet{wang-2020-coding} propose a coarse-to-fine \ac{ICD} path generation framework to generate \ac{ICD} codes from lower levels to higher levels in the \ac{ICD} hierarchy.

Very few publications pay specific attention to rare \ac{ICD} codes and make an effort to improve the prediction performance on rare \ac{ICD} codes.
\citet{DBLP:conf/bionlp/ZhangHZL17} propose a method that retrieves PubMed articles with \ac{ICD} descriptions to enrich the training data of rare \ac{ICD} codes; they extract 500 features from \acp{EHR} and use several classic classifiers, such as SVM and LR, to predict \ac{ICD} codes for each \ac{EHR}.
\citet{frontiers/Teng2020} use the \ac{ICD} description as a query to match content pages in Freebase; the entities extracted from the matched Freebase pages form a knowledge graph, and a better \ac{ICD} representation is learned based on this knowledge graph; finally, they predict \ac{ICD} codes by matching the \ac{EHR} representation with each \ac{ICD} representation. 

Unlike most of the \ac{EHR} coding methods listed above, we aim to improve the prediction of \emph{frequent} and \emph{rare} \ac{ICD} codes at the same time.
Compared with existing methods for improving \emph{frequent and rare} \ac{ICD} code prediction, \OurMethod{}, the graph-based few-shot framework that we propose, can inject external knowledge into the \ac{HEWE} Graphs to enrich \ac{EHR} information.
Moreover, it transfers internal knowledge from predicting frequent \acp{ICD} to rare \ac{ICD} codes in order to improve the prediction performance on rare \ac{ICD} codes.

\subsection{Few-shot learning}
\Ac{FSL} aims to transfer knowledge from frequent classes that generalizes well to classes where only a few samples are available~\cite{DBLP:conf/acl/MadottoLWF19,DBLP:journals/corr/abs-2008-09892}. 
Existing \ac{FSL} methods can be divided into optimization-based methods and metric-based methods.

Optimization-based methods achieve \ac{FSL} by improving the optimization processes, e.g., model initialization, parameter updating. 
E.g., the LSTM-based meta-learner proposed by \citeauthor{iclr/RaviL17} learns the exact optimization algorithm used to train another neural network classifiers in the few-shot regime.
\citet{DBLP:conf/icml/FinnAL17} learns better parameter initialization that is suitable for different \ac{FSL} tasks and is compatible with any model trained with gradient descent.
\citet{DBLP:journals/corr/LiZCL17} present Meta-SGD that learns the parameter initialization, gradient update direction, and learning rate with a one step update.
\citet{DBLP:conf/iclr/MishraR0A18} combine temporal convolution and soft attention to learn an optimal learning strategy.

Metric-based methods are based on nearest neighbor-based methods and kernel density estimation~\citep{DBLP:conf/acl/GengLLSZ20},
which learn an effective metric and similarity function.
Matching networks~\citep{DBLP:conf/nips/VinyalsBLKW16} make predictions by comparing the input example with a few-shot labeled support set by the cosine distance, and weighted the labels of support set to get the prediction label.
\citet{DBLP:conf/nips/SnellSZ17} propose prototypical networks that learn a metric space where classification can be performed by computing squared Euclidean distances to prototype representations of each class.
Unlike fixed metric measures, relation networks learn a deep distance metric to compare the query with given examples~\citep{DBLP:conf/cvpr/SungYZXTH18}.

\OurMethod{} is a metric-based method. 
The differences between the work listed above and our work are least two-fold.
First, we are the first to propose a few-shot \ac{EHR} coding method based on \ac{HEWE} graphs.
Second, we design two graph contrastive learning schemes to alleviate the issue that metric-based methods might fail to encode features that are critical for rare \ac{ICD} codes.

\subsection{Graph contrastive learning}
Contrastive learning is a learning paradigm to learn distinctiveness, i.e., what makes two objects similar or different.
It has received interest due to its success in self-supervised representation learning in natural language processing~\cite{NIPS2013_9aa42b31,DBLP:journals/corr/abs-2104-08821} and computer vision ~\cite{DBLP:conf/cvpr/WuXYL18,DBLP:journals/corr/abs-1906-05849,DBLP:conf/cvpr/He0WXG20}.
Graph contrastive learning extends the paradigm to representation learning on graphs.
\citet{DBLP:journals/corr/abs-2003-01604} learn node representations by randomly selecting pairs of nodes in a graph and training a neural network to predict the contextual position of one node relative to others.
\citet{DBLP:conf/iclr/VelickovicFHLBH19} train a node encoder that maximizes the mutual information between a node representation and the pooled global representation.
\citet{DBLP:journals/corr/abs-2009-05923} define four basic graph alteration operations including edge deletion, edge insertion, node deletion, and node insertion on original graphs to create augmented graphs; 
they then devise a contrastive learning method to distinguish whether two augmented graphs are from the same original graph.
\citet{DBLP:conf/kdd/QiuCDZYDWT20} introduce \ac{GCC}, which defines the pre-training task as subgraph instance discrimination and leverages contrastive learning to learn structural representations; 
their work is the most similar to the \acf{GSCL} that we propose in this work, but \ac{GCC} learns graph structure by determining whether two sub-graphs originate from the same source graph, while \ac{GSCL} captures the intra-correlation in the \ac{HEWE} graph structure by matching the embeddings of two sub-graphs.
In addition, \ac{GSCL} is not independent, and it needs to combine with \ac{GECL} to pre-train the graph encoder.
\ac{GECL} is able to capture invariant, slowly changing features by mining evolving \ac{HEWE} graph sequences.

%% file: 03-preliminaries.tex

\section{Preliminaries}
\subsection{Graph construction}
\begin{figure}[t]
\centering  
\includegraphics[width=1\columnwidth]{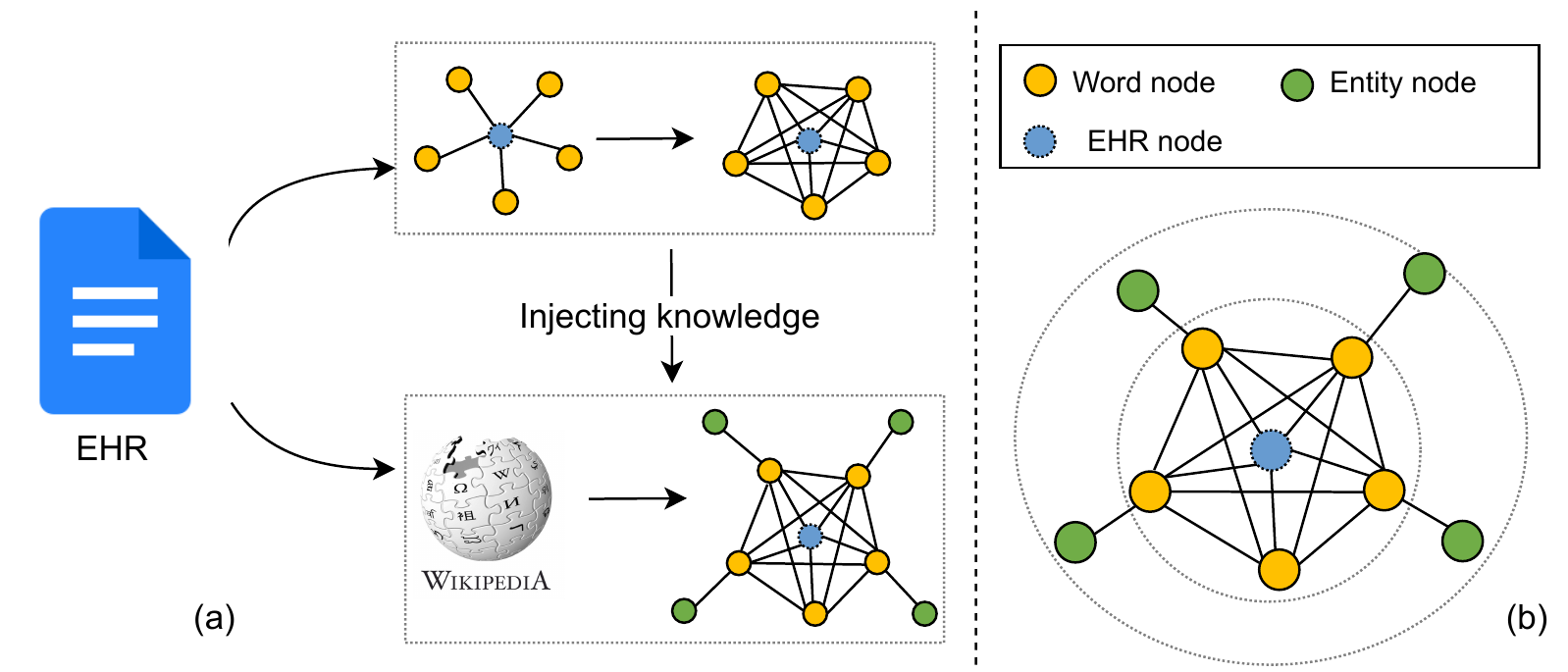}  
\caption{\ac{HEWE} graph construction.}
\label{graph_construction}
\end{figure}
For each \ac{EHR}, we build a \ac{HEWE} graph that can be represented as a tuple $G =(V, E, X)$, where $V$ denotes the set of nodes ${v_1,v_2, \ldots ,v_n}$ and $E$ denotes the set of edges ${e_1, e_2, \ldots, e_m}$.
A \ac{HEWE} graph contains three kinds of nodes: 
\begin{enumerate*}
\item a virtual \ac{EHR} node, 
\item word nodes, and 
\item entity nodes.
\end{enumerate*}
The process for constructing the \ac{HEWE} graph is illustrated in Fig.~\ref{graph_construction}.
First of all, we only preserve high-frequency words in one \ac{EHR} and regard them as word nodes~\cite{10.1145/3357384.3357965}.
An edge between the \ac{EHR} node and a word node is established if the word appears in the corresponding \ac{EHR}.
Then we use a fixed-size sliding window on each \ac{EHR} to gather co-occurrence words, to utilize global word co-occurrence information.
If two words appear in the same sliding window, we will create an edge between the two word nodes.
Finally, we inject the knowledge from Wikipedia\footnote{\url{https://en.wikipedia.org/wiki/Main_Page}} into the \ac{HEWE} graph to enrich the \ac{EHR} information. 
Specifically, we use each word as a query to search the Wikipedia and consider the most relevant entity as the entity node.
Similarly, an edge is established between the word and the entity nodes.
All the edge weights are set to 1.

Moreover, each node $v_i$ is associated with a feature vector $\boldsymbol x_i \in R^{1\times d}$ and $ \boldsymbol X = [x_1; x_2; \ldots ; x_n] \in R^{n \times d}$ denotes all the node features. 
$d$ is the dimension of feature vector and $n$ is the number of all nodes in a \ac{HEWE} graph.
Thus, a \ac{HEWE} graph can be represented as a pair $G = (\boldsymbol A, \boldsymbol X)$, where $\boldsymbol A \in R^{n \times n}$ is an adjacency matrix representing the network structure.
Specifically, $\boldsymbol{A}_{i, j} = 1$ indicates that there is an edge between node $v_i$ and node $v_j$; otherwise, $\boldsymbol{A}_{i, j} = 0$.

\subsection{Problem statement}
Following the common setting in \ac{FSL}, at each episode in the meta-training,
the model takes $C$ \ac{ICD} codes and selects $K$ \ac{HEWE} graphs per \ac{ICD} code to constitute the support set $S$, this problem is also taken as $C$-way $K$-shot \ac{HEWE} graph classification problem.
Formally, given a query \ac{HEWE} graph $q$ and its few-shot support \ac{HEWE} graphs $S$, the task is to design an \ac{EHR} coding model that predicts appropriate \ac{ICD} code for the query \ac{HEWE} graph $q$.
In essence, the objective of this problem is to learn a meta-classifier that can adapt to new \ac{ICD} classification with only a few labeled \ac{HEWE} graphs.
Therefore, how to extract transferable meta-knowledge from training \ac{ICD} codes and transfer the knowledge to testing \ac{ICD} codes is the key for solving the \ac{EHR} coding problem.

%% file: 04-method.tex

\section{Method}
\subsection{Overview}
\begin{figure*}[h]
\centering  
\includegraphics[width=1.05\textwidth]{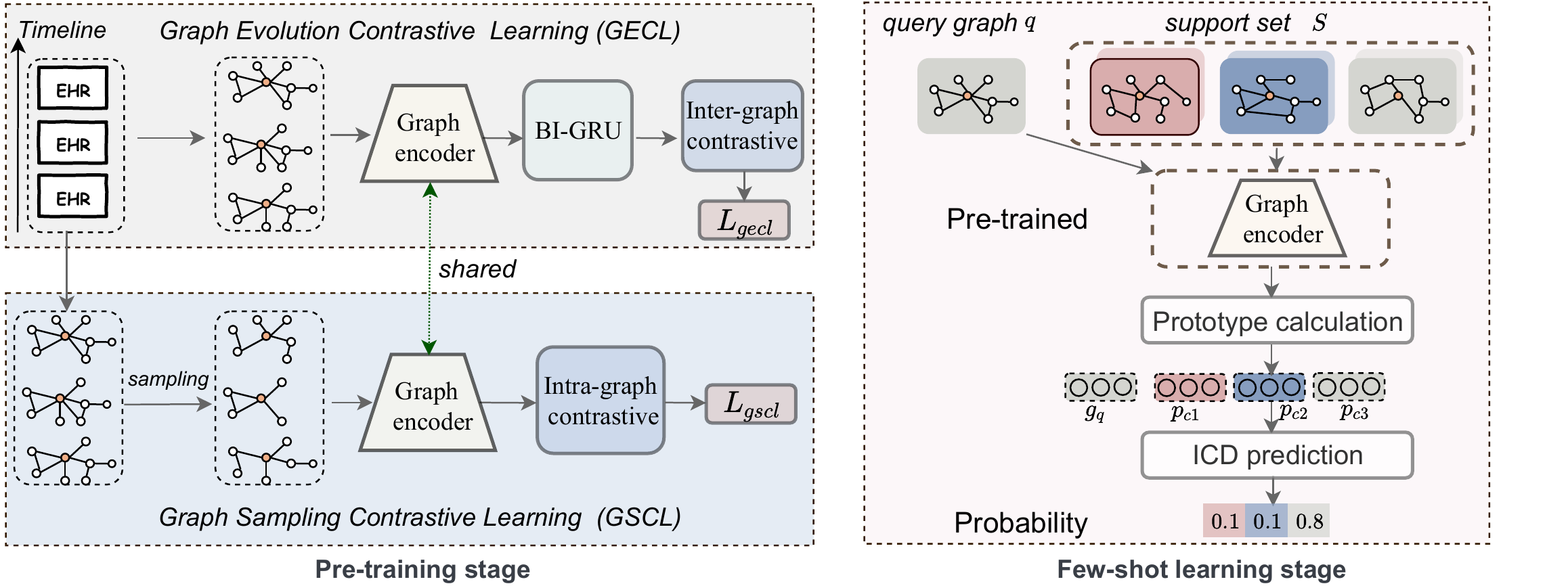}  
\caption{A schematic diagram of proposed \OurMethod{} model.}
\label{model}
\end{figure*}
We elaborate the proposed \OurMethod{} whose framework is illustrated in Figure \ref{model}.
\OurMethod{} consists of two stages: a pre-training stage and a few-shot learning stage.

In the pre-training stage, we pre-train a high-quality graph encoder that is used to initialize the graph encoder in the few-shot learning stage.
The pre-training stage contains two schemes: \acf{GSCL} and \acf{GECL}.
\ac{GSCL} and \ac{GECL} can train a better graph encoder by exploiting graph structure information at the intra-graph level and inter-graph level, respectively.
For the \ac{HEWE} graph sequence of a patient, \ac{GSCL} captures the intra-correlations of graph structure among sampled sub-graphs by intra-graph contrastive learning. 
At the inter-graph level, \ac{GECL} can force the graph encoder to learn the correlations between \ac{HEWE} graphs at different clinical stages by the inter-graph contrastive learning.

The few-shot learning stage is used to predict the most probable \ac{ICD} code for each query \ac{HEWE} graph.
Specifically, as shown in Figure \ref{model}, the support set is composed of several \ac{HEWE} graphs sampled from three different \ac{ICD} codes, i.e., $c_1$, $c_2$, $c_3$, respectively.
First, the query graph and support graphs from the support set are fed into the pre-trained graph encoder to get the representation of each graph.
Second, the prototype calculation module is used to calculate the prototype of each \ac{ICD} code, denoted as $\boldsymbol{p}_{c1}$, $\boldsymbol{p}_{c2}$, and $\boldsymbol{p}_{c3}$.
Third, \OurMethod{} calculates the similarity between the query graph $\boldsymbol{g}_q$ and each \ac{ICD} prototype $\boldsymbol{p}_{c_i}$ ($i=1,2,3$). 
The \ac{ICD} prototype with the highest similarity score ($c_3$ in Figure~\ref{model}, for instance) is the predicted \ac{ICD} for the query graph.

\subsection{Graph sampling contrastive learning}
In order to capture the intra-graph structure and overcome the dependence on \ac{ICD} codes, we introduce the concept of graph sampling.
Formally, given a \ac{HEWE} graph, $G=(\boldsymbol{A},\boldsymbol{X})$, graph sampling is a sampling strategy on nodes and relations of $G$ to produce a random sub-graph $\tilde{G}=(\boldsymbol{\tilde{A}},\boldsymbol{\tilde{X}})$.
There are various sampling strategies for graph data, such as feature corruption and node masking ~\cite{jiao2020sub,DBLP:conf/www/JehW03,DBLP:journals/corr/abs-2001-05140}.
We adopt a node masking-based sub-graph sampling strategy to produce sub-graphs for its simplicity and effectiveness.
Specifically, we regard the \ac{EHR} node in the \ac{HEWE} graph as the central node and randomly mask half of neighbor nodes of the central node.
Correspondingly, the edges associated with these masked nodes are set to 0.
The graphs obtained by node masking are called the \emph{sub-graphs}.
We perform two times of node masking operations randomly to produce two sub-graphs for each \ac{HEWE} graph.

Given one sub-graph $\tilde{G_i}$, a \ac{HEWE} graph encoder is used to encode them to obtain the latent representations matrices $\boldsymbol{\tilde{H_i}}$. 
The corresponding calculation formula can be denoted as:
\begin{equation}
\begin{split}
   \boldsymbol{\tilde{H_i}} & = \text{encoder}(\boldsymbol{\tilde{X_i}}, \boldsymbol{\tilde{A_i}}),\\
\end{split}
\label{encoder}
\end{equation}
where $ \boldsymbol {\tilde{X}_i}$ and $\boldsymbol{\tilde{A}_i}$ represent node feature vector and  adjacency matrix of sub-graph $\tilde{G}_i$ sampled from \ac{HEWE} graph $G_i$.

Here we adopt a two layers \ac{GCN}~\cite{DBLP:conf/iclr/KipfW17}, a flexible node embedding architecture, as the graph encoder.
More specifically, the two layers \ac{GCN} we use can be defined as:

\begin{equation}
\begin{split}
    \boldsymbol{\bar{A_i}} & = \boldsymbol D^{-\frac{1}{2}}\boldsymbol {\breve{A_i}} \boldsymbol D^{-\frac{1}{2}},\\
    \boldsymbol{\tilde{L_i}} & = \text{ReLU}(\boldsymbol {\bar{A_i}}(\boldsymbol{\tilde{X_i}}\boldsymbol{W}_0+\boldsymbol{b}_0)),\\
    \boldsymbol {\tilde{H_i}} & = \text{ReLU}(\boldsymbol{\bar{A_i}}(\boldsymbol \tilde{L_1}\boldsymbol W_1+\boldsymbol b_1)).
\end{split}
\label{gcn}
\end{equation}
Here, $\boldsymbol{\breve{A}}_i={\boldsymbol \tilde{A}}_i+ \boldsymbol{I}_N$ is 
the adjacency matrix of the sub-graph $\tilde{G}_i$ with added self-connections.
$I_N$ is the identity matrix and $D_{i,i}=\sum_{j}\tilde{A}_{i,j}$.
$\boldsymbol{\tilde{X_i}}$ is the initial feature matrix of all nodes in the sub-graph $\tilde{G}_i$;
$\boldsymbol {\tilde{L_i}}$ and $\boldsymbol{\tilde{H_i}}$ are the node representations at different layers in \ac{GCN};
$\boldsymbol{W}_0$ and $\boldsymbol{W}_1$ are parameter matrices;
$\boldsymbol{b}_0$ and $\boldsymbol{b}_1$ are biases.

The central node embedding $\boldsymbol{\tilde{g_i}}$ in sub-graph $\tilde{G}_i$ is picked from the latent representation matrix $\boldsymbol{\tilde{H_i}}$:

\begin{equation}
\begin{split}
   \boldsymbol{\tilde{g_i}} = \mathcal P (\boldsymbol{\tilde{H_i}}), \\
\end{split}
\end{equation}
where $\mathcal P $ denotes the operation of picking out the central \ac{EHR} node embedding from embedding matrix.
Since $\boldsymbol{\tilde{g_i}}$ denotes the central \ac{EHR} node in the sub-graph $\tilde{G}_i$, we can use it as the global representation of sub-graph $\tilde{G}_i$.

We randomly sample a mini-batch of $N$ \ac{HEWE} graphs and produce $2N$ sub-graphs since we sample two times for each \ac{HEWE} graph.
Then we define the \ac{GSCL} task on pairs of sampled sub-graphs derived from the mini-batch.
For a sub-graph sampled from a specific central \ac{EHR} node, our approach for learning the graph encoder relies on contrasting its real sub-graph instances with fake ones.
Specifically, we regard the pair of two sub-graphs sampled from the same \ac{HEWE} graph as a positive pair. 
We do not sample negative examples explicitly.
Instead, given a positive sub-graph pair, similar to \cite{DBLP:conf/kdd/ChenSSH17}, we treat the other $2(N-1)$ sub-graph pairs within a mini-batch
as negative samples.



The loss function we used is termed as the \textit{NT-Xent} ( the normalized temperature-scaled cross entropy loss) and has been used widely in previous work \cite{DBLP:conf/nips/Sohn16,DBLP:conf/cvpr/WuXYL18,DBLP:journals/corr/abs-1807-03748}.
Let $sim(\boldsymbol u,\boldsymbol v)=\boldsymbol u^T \boldsymbol v/\left \| \boldsymbol u \right \|\left \| \boldsymbol v \right \|$ denotes the dot product between $\ell_2$ normalized $\boldsymbol u$ and $\boldsymbol v$ (i.e., cosine similarity).
Then the \emph{NT-Xent} loss for a positive sub-graph pair of $(i, j)$ is defined as:

\begin{equation}
\begin{split}
l_({i,j})=-log\frac{exp(sim(\boldsymbol{g}_i,\boldsymbol{g}_j)/\tau )}{\sum_{k=1}^{2N}\mathbbm{1}_{[k\neq i]}exp(sim(\boldsymbol{z}_i,\boldsymbol{z}_k)/\tau )},
\end{split}   
\label{gscl_loss}
\end{equation}
%
where $\mathbbm{1}_{[k\neq i]} \in \{0,1\}$ is an indicator function evaluating to 1 if $k \neq i$ and $\tau$ denotes a temperature parameter.
In the experiment, we set $\tau$ to 0.5.
The final loss is computed across all positive pairs, both $(i,j)$ and $(j,i)$, in a mini-batch.
Using the pairwise similarities for each positive pair of sub-graphs, the loss function of \ac{GSCL} scheme is aggregated with the following formula:
\begin{equation}
\begin{split}
L_{gscl}=\frac{1}{2N}\sum_{k}^{N}[l(2k-1,2k)+l(2k,2k-1)],
\end{split}   
\label{gscl_loss}
\end{equation}

\subsection{Graph evolution contrastive learning}
An important intuition that we build is that the past states of a patient contain relevant information about their future states. 
Correspondingly, by contrastive learning of past \ac{HEWE} graphs and future \ac{HEWE} graph of each patient, the relevant information between the graphs in the sequence can be captured.
Formally, given a patient $o$, their $T$ \acp{EHR} form an \ac{HEWE} graph sequence according to the recording time.
That is, $o=[G_1,G_2,\ldots ,G_t,\ldots, G_T]$, where $G_t$ represents the $t$-th \ac{HEWE} graph.
Graph evolution is a scheme to learn trends from a history of \ac{HEWE} graphs to predict a future graph.
More specifically, given the historical sequence $[G_1,G_2,\ldots ,G_{T-1}]$, we aim to predict graph $G_T$ at the latest time.

A \ac{BI-GRU} network \cite{DBLP:conf/aipr2/YuZW19} is adopted to model the \ac{HEWE} graph sequence; it is composed of a forward \ac{GRU} unit and a backward \ac{GRU} unit. 
The hidden layer output of the forward \ac{GRU} unit is denoted as 
$\overrightarrow {\boldsymbol h_t}$ and the hidden layer output of the backward \ac{GRU} unit is denoted as $\overleftarrow{\boldsymbol h_t}$. 
The hidden output of a \ac{BI-GRU} at time $t$ is spliced through the hidden layer output of forward \ac{GRU} unit and backward \ac{GRU} unit, as shown in Eq.~\ref{BI-GRU}: 

\begin{equation}
\begin{split}
\overrightarrow{\boldsymbol h_t} & = \text{GRU}(\boldsymbol g_t,\overrightarrow{\boldsymbol h_{t-1}}) \\
\overleftarrow{\boldsymbol h_t} & = \text{GRU}(\boldsymbol g_t,\overleftarrow{\boldsymbol h_{t-1}}) \\
\boldsymbol h_t & =[\overrightarrow{\boldsymbol h_t},\overleftarrow{\boldsymbol h_t}],
\end{split}
\label{BI-GRU}
\end{equation}
where $\boldsymbol h_{t-1}$ is the hidden vector at timestamp $t-1$; $\boldsymbol g_t$ is the representation of $t$-th \ac{HEWE} graph. 
We use the graph encoder shared with \ac{GSCL}  to obtain the representation $\boldsymbol{g}_t$ of $t$-th \ac{HEWE} graph:

\begin{equation}
\begin{split}
    \boldsymbol H_t  & = \text{encoder}(\boldsymbol X_t,\boldsymbol A_t),\\
    \boldsymbol{g}_t & = \mathcal P (\boldsymbol{H}_t), 
\end{split}
\end{equation}
where $\boldsymbol X_t$ denotes node features and $\boldsymbol A_t$ denotes the adjacency matrix in the $t$-th \ac{HEWE} graph.
We use $\boldsymbol h_{T-1}$ at timestamp $T-1$ as the  context representation that forms a summary of past \ac{HEWE} graph sequence.

Similar to \ac{GSCL}, we randomly sample a mini-batch of $N$ \ac{HEWE} graph sequences and then define the \ac{GECL} task on $2N$ samples derived from the mini-batch.
For a \ac{HEWE} graph sequence, our approach for learning the graph encoder relies on contrasting its real coherent sequence with fake ones.
Specifically, the context representation $\boldsymbol h_{T-1}$ and future graph $\boldsymbol g_T$ then form a \emph{positive pair}, while replacing the future graph $\boldsymbol g_T$  with  $\boldsymbol g_T*$ from other \ac{HEWE} graph sequences will generate \emph{negative pairs}.  



%

Given a pair $(\boldsymbol h_{T-1}$, $\boldsymbol g_T)$, a powerful mechanism to incorporate the interaction between them is via bilinear pooling~\cite{fukui2016multimodal}. 
This is essentially an outer product and results in a matrix $\boldsymbol Z \in \mathbb{R}^{d\times d}$:

\begin{equation}
\boldsymbol Z=\boldsymbol h_{T-1} \otimes \boldsymbol g_T=\boldsymbol{h}_{T-1} \times \boldsymbol g_T^{T},
\end{equation}
where $\otimes$ denotes the outer product operation of two emebddings.
This is followed by a reshaping of the matrix $\boldsymbol Z$ into $\boldsymbol z \in \mathbb{R}^{d^2 \times 1}$:

\begin{equation}
\boldsymbol z = \text{reshape}(\boldsymbol Z),  
\end{equation}
which, in turn, is followed by a linear projection of the result to obtain the output of the interaction $ u $:

\begin{equation}
  u = \boldsymbol W_u \times \boldsymbol z,
\end{equation}
where the $\boldsymbol W_u$ is a learned transformation matrix. 
The objective of \ac{GECL} is to predict whether the candidate graph $\boldsymbol g_T$ is a correct future graph for the patient.
We use the binary cross-entropy loss for model optimization:

\begin{equation}
    L_{gecl}=-\frac{1}{N^2}\sum_{i=1}^{N^2}y_i\cdot \log(\sigma (u)+(1-y_i)\cdot \log(1-\sigma(u)),
\end{equation}
where $\sigma(u)=1/(1+\exp(-u))$ is the sigmoid function and $y_i$ denotes whether the $i$-th context-future pair is positive or negative, and $M$ denotes the total number of context-future pairs.

\subsection{Co-training}
The two graph contrastive learning schemes, i.e., \ac{GSCL} and \ac{GECL}, do not depend on \ac{ICD} labels.
So we can easily extend them to learn as well from additional unlabeled \ac{HEWE} graphs and we can make \ac{GSCL} and \ac{GECL} benefit from each other by jointly training:

\begin{equation}
    \min_{\theta } \left(L_{gecl}+\alpha L_{gscl}\right),
\label{gecl-gscl-loss}
\end{equation}
where $\alpha$ controls the importance of the \ac{GSCL} task and $\theta$ represents all learnable parameters. 
In the experiments, we set $\alpha$ to 0.5.

\subsection{Graph few-shot \ac{EHR} coding}
\subsubsection{Overall architecture of few-shot \ac{EHR} coding}
In the few-shot learning stage, we construct an ``episode'' to compute gradients and update our model in each training iteration\cite{DBLP:journals/corr/abs-2004-05805}. 
For the $C$-way $K$-shot problem, a training episode is formed by randomly selecting $C$ \ac{ICD} codes from the $C_{train}$ (frequent \ac{ICD} codes) and choosing $K$ \ac{HEWE} graphs per selected \ac{ICD} to act as the support set $S= \bigcup_{i=1}^{C}\{G_{c_i,s},y_{c_i,s}\}_{s=1}^{K}$.
A subset of the remaining examples serves as the query set $ Q=\{G_q,y_q\}_{q=1}^L$. 
Training on each episode is conducted by feeding the support set $S$ to the model and updating its parameters to minimize the loss in the query set $Q$. 

\subsubsection{Pre-trained graph encoder}
In order to learn an expressive \ac{HEWE} graph representation for each \ac{EHR}, we develop a graph encoder to capture the graph structure.
The encoder structure is exactly the same as the graph encoder in \ac{GSCL} and \ac{GECL}, which is a two-layer \ac{GCN}. 
Please refer to Eq.~\ref{gcn} for details.

\begin{equation}
\begin{split}
    \boldsymbol{H}_{c_i,j} & =\text{encoder}(\boldsymbol X_{c_i,j},\boldsymbol A_{c_i,j}),\\
     \boldsymbol{g}_{c_i,j} & = \mathcal P (\boldsymbol{H}_{c_i,j}), 
\end{split}
\label{fsl-encoder}
\end{equation}
where $\boldsymbol{g}_{c_i,j}$ represents the $j$-th \ac{HEWE} graph and it belongs to \ac{ICD} $c_i$ in the support set $S$;
$\boldsymbol{X}_{c_i,j}$ and $\boldsymbol{A}_{c_i,j}$ represent the node features and adjacency matrix corresponding to this \ac{HEWE} graph, respectively.

We use \ac{GSCL} and \ac{GECL} as tasks to pre-train a \ac{HEWE} graph encoder and use it to initialize the graph encoder in the few-shot \ac{EHR} learning stage.

\subsubsection{Prototype calculation}
\begin{figure}[t]
\centering  
\includegraphics[width=0.5\columnwidth]{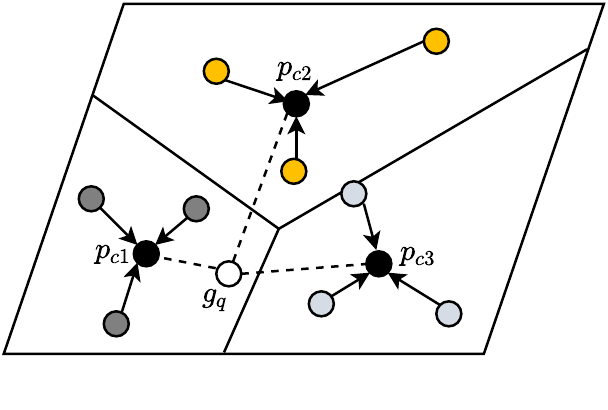}  
\caption{Prototypical Network for few-shot learning}
\label{fig_prototype}
\end{figure}
We follow the idea of prototypical net\-works \cite{DBLP:conf/nips/SnellSZ17}, which encourages \ac{HEWE} graphs in the support set $S_{c_i}$ to cluster around a specific prototype.
The schematic diagram of the prototypical network is illustrated in Fig~\ref{fig_prototype}. 
In this figure, $\boldsymbol{p}_{c_1}$, $\boldsymbol{p}_{c_2}$ and $\boldsymbol{p}_{c_3}$ denote the prototype representation for \ac{ICD} code $c_1$, $c_2$ and $c_3$, respectively.
Formally, the prototype representation of \ac{ICD} $c_i$ can be computed by:
\begin{equation}
    \boldsymbol p_{c_i}=P_\mathit{roto}(\{\boldsymbol g_{c_i,j}|\forall j \in S_{c_i}\}),
\end{equation}
where $S_{c_i}$ denotes the set of labeled \ac{HEWE} graphs from \ac{ICD} code $c_i$ and $P_\mathit{roto}$ is the prototype computation function;
$\boldsymbol g_{c_i,j}$ is the $j$-th \ac{HEWE} graph representation for the \ac{ICD} $c_i$.
The prototype representation $\boldsymbol p_{c_i}$ of the \ac{ICD}  $c_i$ is computed by taking the average of all embedded \ac{HEWE} graphs belonging to the \ac{ICD} $c_i$:

\begin{equation}
    \boldsymbol{p}_{c_i}=\frac{1}{|S_{c_i}|}\sum_{j=1}^{|S_{c_i}|}g_{c_i,j}
\label{prototype}
\end{equation}

\subsubsection{\ac{ICD} prediction}
The learned prototypes define a \ac{ICD} predictor for each query \ac{HEWE} graph $\boldsymbol{g}_q$.
More specifically, it can assign a probability over \ac{ICD} code $c_i$ based on the matching score between the query \ac{HEWE} graph $\boldsymbol{g}_q$ and each \ac{ICD} prototype $\boldsymbol{p}_{c_i}$.
Specifically, we model this matching relationship as follows:

\begin{equation}
   p(c_i|\boldsymbol{g}_q)= \sigma (\boldsymbol W_{c}(\boldsymbol g_q \parallel \boldsymbol p_{c_i})),
\label{prediction}
\end{equation}
where $ \parallel$ denotes the concatenation operation, and $\boldsymbol W_{c}$ is the weight matrix; 
$\boldsymbol g_q$ is the query \ac{HEWE} graph representation and $\boldsymbol p_{c_i}$ is the prototype representation of class $c_i$;
$p(c_i|g_q)$ is the probability of the query \ac{HEWE} graph belonging to \ac{ICD} $c_i$.

Under the episodic training framework, the objective of each episode is to minimize the classification loss between the predictions of the query set and the ground-truth.
Specifically, for each episode, given the support set $S$ and query set $Q=\{G_q,y_q\}_{q=1}^L$, the training objective is to minimize the cross-entropy loss on the $C$ \ac{ICD} codes:
\begin{equation}
   L_\mathit{few}(S,Q)=-\frac{1}{C}\sum_{i=1}^{C}\frac{1}{L}\sum_{q=1}^{L}y_q \log(p(c_i|\boldsymbol{g}_q)).  
\label{query_loss}
\end{equation}
By minimizing the above loss function, \OurMethod{} is able to learn a generic classifier for the $C_\mathit{train}$ \ac{ICD} codes.
After training on a considerable number of episodes, its generalization performance will be measured on the test episodes, which contain \ac{HEWE} graphs sampled from $C_\mathit{test}$ (rare \ac{ICD} codes) instead of $C_\mathit{train}$.
For each test episode, we use the predictor produced by \OurMethod{} for the provided support set $S$ to classify each query \ac{HEWE} graph in $Q$.
The details of training and testing the \OurMethod{} are shown in Algorithm \ref{alg:fsl}.
Firstly, the graph encoder is pre-trained by the \ac{GECL} and \ac{GSCL} schemes (line 2-3).
Next, the \ac{EHR} coding model is trained using the labelled data (line 4-10).
Finally, the \OurMethod{} is tested for each test episode (line 12-18).

\begin{algorithm}[t]
\SetAlgoNoLine
\SetKwInOut{Input}{\textbf{Input}}\SetKwInOut{Output}{\textbf{Output}}
\BlankLine 
\textcolor[rgb]{0.3,0.3,0.3}{//\emph{Training process}}\\
Jointly train \ac{GSCL} and \ac{GECL} by minimizing the loss in Eq.\ref{gecl-gscl-loss};\\
Save the parameters of the graph encoder in Eq.\ref{encoder};\\
Initialize the graph encoder in Eq.\ref{fsl-encoder} with the parameters saved in line 3;\\
\For{each episode iteration}{

    Randomly select $C$ \ac{ICD} codes from the $C_{train}$;\\
    Randomly select $K$ labeled \ac{HEWE} graphs from each of the $C$ \acp{ICD} as support set $S$;\\
    Randomly select a fraction of the reminder of those $C$ \ac{ICD} codes’ \ac{HEWE} graphs as query set $Q$;\\
      
    Feed  $S$ to the model and minimizing the loss in Eq.\ref{query_loss};\\
    } 
    
\textcolor[rgb]{0.3,0.3,0.3}{\emph{//Testing process}}\\  
\For{each episode iteration}{

    Randomly select $C$ \ac{ICD} codes from the $C_{test}$;\\
    Randomly select $K$ labeled \ac{HEWE} graphs from each of the $C$ \acp{ICD} as support set $S$;\\
    Randomly select a fraction of the reminder of those $C$ \ac{ICD} codes’ \ac{HEWE} graphs as query set $Q$;\\
      
    Calculate the prototypes for $C$ \ac{ICD} codes with Eq.\ref{prototype};\\
    Predict \ac{ICD} for each \ac{HEWE} graph in $Q$ with Eq.\ref{prediction}.
    } 

\caption{The training and testing process of \OurMethod{}}
\label{alg:fsl}
\end{algorithm}

%% file: 05-evaluation.tex

\section{Experimental Setup}
\subsection{Dataset}
We conduct experiments on the MIMIC-III dataset\footnote{The dataset used in this paper is available at \url{https://mimic.physionet.org/}.}, which is a large, freely-available database comprising de-identified health-related data associated with over forty thousand patients between 2001 and 2012. 
This is the only benchmark dataset that is commonly used on this task and publicly available~\cite{johnson2016mimic}.
As with previous studies, e.g.,\cite{DBLP:conf/naacl/MullenbachWDSE18,aaai/PrakashZHDLQLF17}, we focus on discharge summaries in \ac{EHR}, which summarize the information about a stay into a single document.
%
We count the number of occurrences of each \ac{ICD} code in the MIMIC-III dataset, and \ac{ICD} codes are selected as training classes if they occur more than 20 times. 
Among the remaining \ac{ICD} codes, the number occurrences of \ac{ICD} codes selected as testing class should be no less than 2 times, and the remaining \ac{ICD} codes are filtered. 
It should be noted that there is no overlapping between training \ac{ICD} and testing \ac{ICD} codes.
Accordingly, these \acp{EHR} are divided into training set and test set according to their \ac{ICD} codes.
The one more thing is we randomly divide the training dataset into training and validation sets with a 7:3 ratio during training.
Descriptive statistics of the training set and test set are given in Table \ref{tb:dataset}.
It can be observed from Table \ref{tb:dataset} that the number of \ac{ICD} codes in the test set is more than that in the training set, and the average number of \ac{HEWE} graphs in the test set is only 6.43, which is much smaller than that in the training set.
Again, this indicates that the distribution of the \ac{ICD} codes is extremely imbalanced, which demonstrates the challenge of this task.

\begin{table}[t]
\centering
\setlength{\tabcolsep}{3pt}
\caption{Statistics of the MIMIC-III dataset.}
\label{tb:dataset} 
\begin{tabular}{l rr}
\toprule
\bf Statistics & \bf Training set  & \bf Test set\\ 
\midrule
\# \ac{ICD} codes  &2,033 & 3,335 \\ 
avg \# of \ac{HEWE} graphs & 284.45      & 6.43 \\ 
\midrule
 \# of \ac{EHR} nodes & 51,434     & 14,519 \\ 
avg \# of word nodes & 117.89      & 101.05 \\ 
avg \# of entity nodes &177.11      & 187.95 \\ 
\midrule
avg \# of \ac{EHR}-word edges &  118.01    & 101.23 \\ 
avg \# of word-word edges &   3,903.71  & 3,293.19   \\ 
avg \# of word-entity edges &  92.85     & 101.45 \\ 
\bottomrule
\end{tabular}
\end{table}

\subsection{Baselines}
In order to demonstrate the effectiveness of \OurMethod{}, we compare it with several methods, including state-of-the-art models for \ac{EHR} coding and few-shot learning methods:

\begin{itemize}[leftmargin=*,nosep]
\item \textbf{\Ac{MLP}}. 
We learn a multi-label classification model with a three-layer perceptron to predict the probability of each \ac{ICD} code.

\item \textbf {BI-GRU}~\cite{acl/ZhouSTQLHX16}.
This method uses a bidirectional gated recurrent unit to encode \ac{EHR} and then performs binary classification on each \ac{ICD} code based on the \ac{EHR} representation.

\item \textbf {CAML} and \textbf{DR-CAML} \cite{naacl/MullenbachWDSE18}.
CAML exploits Text-CNN \cite{emnlp/Kim14} to obtain the representation of each \ac{EHR} and then uses label-depen\-dent attention to learn the most informative representation for each \ac{ICD} code, based on which it does binary classification.
DR-CAML enhances CAML by adding an \ac{ICD} description regularization term to the final classification weights.

\item \textbf {MSATT-KG} \cite{cikm/Xie18}.
MSATT-KG leverages a convolutional neural network to produce variable n-gram features for clinical notes and incorporates multi-scale
feature attention to adaptively select multi-scale features.
The multi-scale features are used to perform multi-label classification over all the \ac{ICD} codes.

\item \textbf {Matching Network (MatchingNet)}~\cite{DBLP:conf/nips/VinyalsBLKW16}.
MatchingNet first gets the representations of support \ac{HEWE} graphs and query \ac{HEWE} graphs, and then computes the similarity of query graph to each support graph. 
Finally, the \acp{ICD} from the support \ac{HEWE} graphs are weight-blended together accordingly to predict \ac{ICD} codes.

\item \textbf {PrototypeNet} and \textbf {PrototypeNet-attention}~\cite{snell2017prototypical}.
PrototypeNet computes the distance between query \ac{HEWE} graph and each \ac{ICD} prototype. 
Then it determines the \ac{ICD} of the query \ac{HEWE} graph based on the shortest distance.
PrototypeNet-attention assigns different weights to each support graph when calculating the prototype representation of each \ac{ICD}.

\item \textbf {RelationNet-attention}~\cite{sung2018learning}.
RelationNet-attention is the RelationNet with an attention mechanism; it is composed of an embedding module and a relation module. 
The embedding module produces the representation of the query and support \ac{HEWE} graphs.
Then the relation module compares these embeddings to determine whether they belong to the same \ac{ICD}.
\end{itemize}

\subsection{Evaluation metrics}
A discussion of performance metrics used for \ac{EHR} coding is beyond the scope of this paper~\cite{aaai/PrakashZHDLQLF17,naacl/MullenbachWDSE18,cikm/Xie18}.
We select macro-averaged metrics, i.e., ACC, Precision, Recall, F1, because they are calculated by averaging metrics computed per-\ac{ICD}~\citep{goutte2005probabilistic, huang2005using}.
ACC refers to the percentage of correctly classified \acp{EHR} among the total number of \acp{EHR}. 
In the case of an imbalanced distribution between different \ac{ICD} codes, accuracy alone is not enough and more appropriate metrics are Precision and Recall.
F1 score is the harmonic mean of Precision and Recall, so it is a relatively impartial evaluation metric.

\subsection{Implementation details}
For \OurMethod{}, the initial embedding dimension of each node is set to 300.
We use a two layer \ac{GCN} to get the embedding representation of each node and the dimensions of $\boldsymbol W_0$ and $\boldsymbol W_1$ (Eq.~\ref{gcn}) are set to 100 and 300, respectively.
In the pre-training stage of the graph encoder, we use a \ac{BI-GRU} to model the graph evolution representation, the hidden size of the \ac{BI-GRU} in Eq.~\ref{BI-GRU} is set to 300,
and the value of $\alpha$  in Eq.\ref{gecl-gscl-loss} is set to 0.5.
We use the Adam optimizer to optimize the parameters of the two graph contrastive learning schemes. 
The batch size is to 128 and the learning rate is 0.0001.
In the few-shot learning stage, the \ac{ICD} number $C$ in each episode is set to 5. 
We pick 5 \ac{HEWE} graphs for each \ac{ICD} code as the support set, and the query set has 15 \ac{HEWE} graphs for each \ac{ICD} code in every training episode.
That is, $K=5$ and $L=15$ when trains \OurMethod{}.
In the testing stage, we use $C=5$ and $K=5$ for comparison.
We initialize the parameters of \OurMethod{} randomly and use a batch size of 64 in the few-shot learning stage. 
The Adam optimizer~\cite{DBLP:journals/corr/KingmaB14} is used to optimize the parameters of \OurMethod{} and the maximum training epochs is set 300.
The learning rate $\lambda$ is 0.001 and the momentum parameters are set to the defaults $\beta1=0.9$ and $\beta2=0.999$.
We implement \OurMethod{} in Pytorch and train it on a GeForce GTX TitanX GPU\footnote{The source code is available at \url{https://github.com/WOW5678/CoGraph}.}.

%% file: 06-results.tex

\section{Result}

\begin{table*}[]
\centering
\caption{Performance comparison (\%) of different methods. \textbf{Bold face} indicates the best result in terms of the corresponding metric.
Significant improvements over the best baseline results are marked with $^\ast$ (t-test, $p < 0.05$).}
\label{tb_result} 
\begin{tabular}{l cc cc cc cc}
\toprule
& \multicolumn{4}{c}{\bf Rare \ac{ICD} codes} & \multicolumn{4}{c}{\bf Frequent \ac{ICD} codes}  
\\
\cmidrule(r){2-5}\cmidrule{6-9}
\bf Method  & ACC & Precision  & Recall & F1  & ACC & Precision  & Recall & F1  
\\
\midrule
\ac{MLP}  & 19.62 & \phantom{0}7.98 & 20.70 & 10.41 & 20.41 & \phantom{0}9.07 & 20.73  & 11.31  \\ 
BI-GRU    & 20.91 & \phantom{0}8.06 & 21.76 & 11.14 & 36.04 & 36.29 & 38.37  & 35.55   \\ 
\midrule 
CAML      & 23.56 & 16.44 & 17.80 & 16.86 & 63.02 & 69.94 & 62.11  & 63.78    \\
DR-CAML   & 16.77 & 17.43 & 17.48 & 16.97 & 62.60 & 68.98 & 62.41  & 63.76   \\ 
MSATT-KG  & 16.04 & 17.28 & 16.58 & 15.44 & 63.33 & 63.53 & 64.79  & 62.18   \\  
\midrule 
MatchingNet  & 54.95 &  60.17 & 54..90 & 54.56 & 61.56 & 68.22 & 61.89  & 61.23  \\
PrototypeNet & 52.21 & 61.34 &  51.54 & 51.96 & 61.04 & 67.09 & 59.70  & 60.97  \\
PrototypeNet-attention & 54.73 & 61.49 & 54.80 & 54.69
&63.96 & \bf 73.20 & 64.22  & 65.60   \\
RelationNet-attention& 52.92 & 55.75 & 53.11 & 53.02 & 61.88 & 68.79 & 61.94  & 62.71   \\
\midrule 
\OurMethod & \bf 57.07\rlap{$^\ast$} & \bf 69.81 & \bf 56.37 & \bf 57.64\rlap{$^\ast$}& \bf 65.27\rlap{$^\ast$} &  71.46 & \bf 65.32\rlap{$^\ast$}  & \bf 66.21\rlap{$^\ast$}  \\
\bottomrule       
\end{tabular}
\end{table*}

\subsection{Performance on rare \ac{ICD} codes}
To assess the performance of \OurMethod{} on rare \ac{ICD} codes (i.e., codes that occur fewer than 10 times, and more than 2 times in the MIMIC-III data), we report the evaluation metrics for rare \ac{ICD} codes in the left half of Table \ref{tb_result}.
Based on Table \ref{tb_result}, we have several observations.
First, \OurMethod{} achieves the best performance on most of the evaluation metrics, such as ACC, Precision, Recall and F1.
This indicates that \OurMethod{} is able to effectively perform rare \ac{ICD} classification by transferring  knowledge between different \ac{ICD} codes.
Importantly, the F1 get an improvement of 2.95\%  over the best baseline, i.e., PrototypeNet-attention, and compared to the worst baseline, i.e., \ac{MLP}, the improvement in F1 is 47.23\%.
The reason for these improvements is that \OurMethod{} can transfer classification knowledge between different \ac{ICD} codes, using our two graph contrastive learning schemes to capture the intra-graph and inter-graph correlations.

Second, on the classification of rare \ac{ICD} codes, the few-shot learning based methods (i.e., MatchingNet, PrototypeNet, Prototype\-Net-attention and RelationNet-attention) perform better than other methods (i.e., \ac{MLP}, \ac{BI-GRU}, CAML, DR-CAML and MSATT-KG).
Among them, \ac{MLP} and \ac{BI-GRU} are relatively simple baselines and perform poorly, while CAML, DR-CAML and MSATT-KG all consider the description information of \ac{ICD} codes, and have more powerful architectures, so compared to \ac{MLP} and \ac{BI-GRU}, they can get better results.
Since the imbalance in the distribution of \ac{ICD} codes and the lack of training samples are not considered by the non-few-shot learning methods, their performance is much worse than that of the few-shot leaning based methods.

Third, the performance differences between different few-shot learning methods are small.
For example, the F1 of Prototype-attention only improves by 0.13\% over the MatchingNet. 
This is because these few-shot learning methods are all metric-based methods, and the differences between them are only in the similarity measurements of support set and query set.
Thanks to its attention mechanism, the accuracy and precision of PrototypeNet-attention have been improved by 2.73\% compared to PrototypeNet.

\subsection{Performance on frequent \ac{ICD} codes}
To assess the performance of \OurMethod{} on frequent \ac{ICD} codes, we report the evaluation metrics for frequent \ac{ICD} codes in the right half of Table \ref{tb_result}.
Based on Table \ref{tb_result}, we have some observations.
First, \OurMethod{} achieves the best performance on most of the evaluation metrics, such as ACC, Recall and F1, but not Precision.
This indicates that \OurMethod{} is able to effectively perform frequent \ac{ICD} classification by episode-based meta-training on different \ac{ICD} codes.
The F1 gets an slight improvement of 0.61\% over the best baseline, i.e., PrototypeNet-attention, and compared to the worst baseline, i.e., \ac{MLP}, the increasement in F1 is 54.90\%.
In addition, the recall values are significantly lower than precision values in most of the baselines and {\OurMethod}, indicating that they tend to neglect some correct \ac{ICD} codes.

Second, unlike on rare \ac{ICD} codes, CAML, DR-CAML and MSATT-KG perform better than some few-shot learning-based methods, such as MatchingNet and PrototypeNet.
These results indicate that CAML, DR-CAML and MSAT-KG methods are very effective if they are fed sufficiently many training samples, but they cannot obtain a good classification model when the training set contains only a few training samples.

Third, compared to frequent \ac{ICD} codes, \OurMethod{} has a bigger advantage over competing methods on rare \ac{ICD} codes. 
For example, on the rare \ac{ICD} codes, the F1 of \OurMethod{} improves by 2.95\% over the best baseline.
On the frequent \ac{ICD} codes, our \OurMethod{} also gets a slight improvement, i.e., 0.61\%, over the best baseline.
The reason is that \OurMethod{} is pre-trained by two effective graph-based contrastive learning schemes, i.e., \ac{GSCL} and \ac{GECL}, which prompt the \OurMethod{} to learn graph features that are independent of the \ac{ICD} codes.
Furthermore, these independent graph features can be transferred to the \ac{ICD} coding task in the few-shot learning stage.

%% file: 07-analysis.tex

\section{Analysis}
\subsection{Ablation study}
\begin{table*}[]
\centering
\caption{Analysis of different components in \OurMethod{}. \textbf{Bold face} indicates the best result in terms of the corresponding metric.
Significant improvements over the best baseline results are marked with $^\ast$ (t-test, $p < 0.05$).}
\label{ablation} 
\begin{tabular}{l cc cc cc cc}
\toprule
& \multicolumn{4}{c}{\bf Rare \ac{ICD} codes} & \multicolumn{4}{c}{\bf Frequent \ac{ICD} codes}  
\\
\cmidrule(r){2-5}\cmidrule(r){6-9}
\bf Method  & ACC & Precision  & Recall & F1  & ACC & Precision  & Recall & F1  
\\
\midrule
-\ac{GECL}-\ac{GSCL}  & 50.30  & 64.72  & 50.73  & 50.37 & 60.62 & 64.63 & 60.07  & 60.09  \\
-\ac{GECL} & 55.90 & 69.16 & 56.15 & 56.36 & 62.71 & \bf 71.64 & \bf 65.91  & 65.27 \\ 
-\ac{GSCL} &56.05  & 69.00 & 55.92 & 56.61 & 63.02 & 68.72 & 61.78  & 63.30   \\\midrule
\OurMethod{} & \bf 57.07 & \bf 69.81 & \bf 56.37\rlap{$^\ast$} & \bf 57.64 & \bf 65.27 &  71.46 &  65.32  & \bf 66.21\rlap{$^\ast$}  \\
\bottomrule       
\end{tabular}
\end{table*}
We conduct an ablation study to analyze the effects of different components in \OurMethod{}, as shown in Table~\ref{ablation}.
These are the variants of our method that we consider:
\begin{enumerate*}
\item \textit{-\ac{GECL}-\ac{GSCL}} denotes \OurMethod{} without graph contrastive learning schemes. 
We remove the graph contrastive learning schemes and just use graph-based few-shot learning to train our model.
\item \textit{-\ac{GECL}} denotes \OurMethod{} without \ac{GECL}.
\item \textit{-\ac{GSCL}} denotes \OurMethod{} without \ac{GSCL}.
\end{enumerate*}
From the results, we obtain the following insights:

First, the performance of \OurMethod{} on both rare and frequent \ac{ICD} codes decreases dramatically after removing graph evolution and graph sampling contrastive leaning (i.e., \textit{-GECL-GSCL}).
Specifically, the F1 values drop by 7.27\% and 6.12\% on rare and frequent \ac{ICD} codes, respectively.
The result shows that using graph contrastive learning schemes to pre-train the graph encoder plays a crucial role in the few-shot learning stage.
This is because \ac{GSCL} and \ac{GSCL} do not require the supervision of \ac{ICD} codes, which helps to learn more general graph features. 
At the same time, \ac{GECL} captures the inter-graph correlations by comparing graph sequences, while \ac{GSCL} contrasts sub-graphs to obtain the intra-graph correlations between them. 
Both contrastive learning schemes can pre-train a more robust graph encoder. 
In the few-shot learning stage, the graph encoder is fine-tuned to adapt to the classification of different \ac{ICD} codes.

Second, compared to frequent \ac{ICD} codes, the \ac{GSCL} and \ac{GECL} schemes have almost the same effect on rare \ac{ICD} codes. 
When the \ac{GECL} and \ac{GSCL} schemes are removed, the model’s ACC and F1 decreases by 1.25\% and 1.03\% on rare \ac{ICD} codes and by 0.94\% and 2.91\% on frequent \ac{ICD} codes.
That's because the graph contrastive learning schemes do not require \ac{ICD} labels.
More specifically, \ac{GSCL} and \ac{GECL} prompt the {\OurMethod} to learn features that do not rely on \ac{ICD} labels.
So compared to rare \ac{ICD} codes, the pre-trained graph encoder plays almost the same role in the frequent \ac{ICD} classification even though the data is unbalanced.
These figures confirms that class-imbalanced learning can significantly benefit in self-supervised manners~\cite{NEURIPS2020_e025b627}.

Third, after removing the \ac{GECL} or \ac{GSCL} module, the performance of \OurMethod{} decreases on both rare \ac{ICD} and frequent \ac{ICD}.
Specifically, on the rare \ac{ICD}, the ACC of \OurMethod{} drops from 57.07\% to 55.90\% (i.e., -\ac{GECL}) and to 56.05\% (i.e.,-\ac{GSCL}), while on the frequent \ac{ICD}, the ACC of \OurMethod{} drops by 2.56\% and 2.25\% respectively.
This fully shows that the graph encoder pre-trained by \ac{GECL} and \ac{GSCL} has learned different aspects of the \ac{HEWE} graph and that \ac{GECL} and \ac{GSCL} are helpful for the classification of rare and frequent \ac{ICD} codes.
This result is in line with our initial intuition: \ac{GECL} learns correlations between graphs, while  \ac{GSCL} learns the intra-graph correlations.
So it is effective for \OurMethod{} to combine \ac{GECL} and \ac{GSCL} to learn graph structures.
\subsection{Graph embedding visualization}
\begin{figure*}[t]
\centering  
\includegraphics[width=2.1\columnwidth]{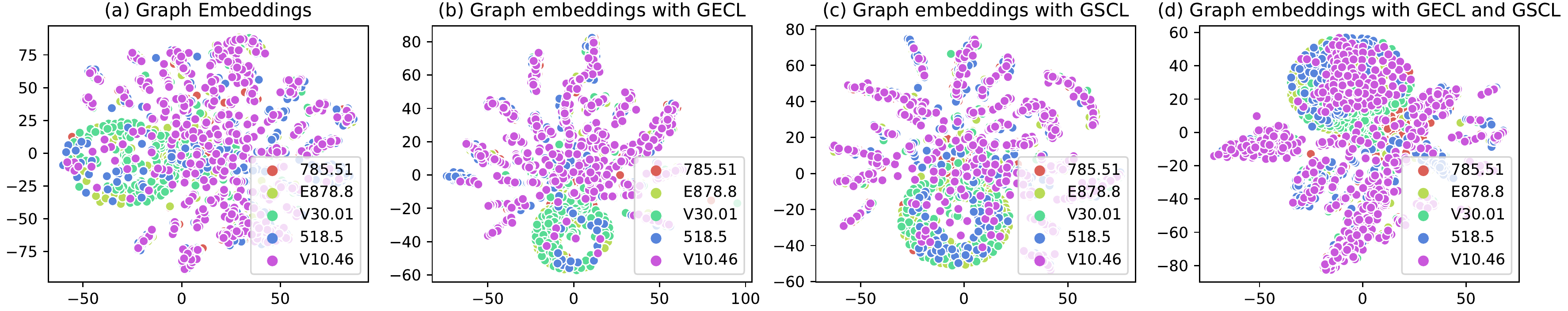}  
\caption{Graph embedding with different pre-trained graph encoder.}
\label{embedding}
\end{figure*}

In this section, we analyze the \ac{HEWE} graph embeddings learned using two contrastive schemes to see if they learn any informative representations during the pre-training stage.
We randomly select 5 \ac{ICD} codes from the test set, each of which contains 1000 samples, and use the t-Distributed Stochastic Neighbor Embedding (t-SNE)~\cite{van2008visualizing} for high-dimensional embedding exploration and visualizing.
The results are shown in Fig.~\ref{embedding}.

Comparing Fig.~\ref{embedding}(d) with the other figures in terms of their densities of \ac{HEWE} graph embeddings, we can see that graph embeddings with both \ac{GSCL} and \ac{GECL} can learn tighter clusters for the \ac{ICD} codes.
For example, most of the samples belonging to V10.46 are closely located to one another in Fig.~\ref{embedding}(d).
This indicates that pre-training the graph encoder by combining \ac{GSCL} and \ac{GECL} can obtain better representations of the \ac{HEWE} graphs, compared with the other cases, i.e,  without pre-training (Fig.~\ref{embedding}(a)) or using a single graph contrastive learning scheme (Fig.~\ref{embedding}(b) and Fig.~\ref{embedding}(c)).

It is worth noting that although most of the samples belonging to the same \ac{ICD} code are grouped together, the boundaries between different clusters are not obvious.
We think that this is related to the goal of few-shot learning, which is to find out \emph{the most similar support samples} for the query samples by maximizing the similarities between query and support samples~\cite{DBLP:conf/igarss/LiZWZ20}, so as to conduct the \ac{ICD} prediction for the query samples.
It does not care about the mapping between the samples and the \ac{ICD} codes.
As a result, there are no clear boundaries among the different \ac{ICD} clusters. 

\subsection{Impact of few-shot size}
\begin{figure}[t]
\centering  
\includegraphics[width=1.0\columnwidth]{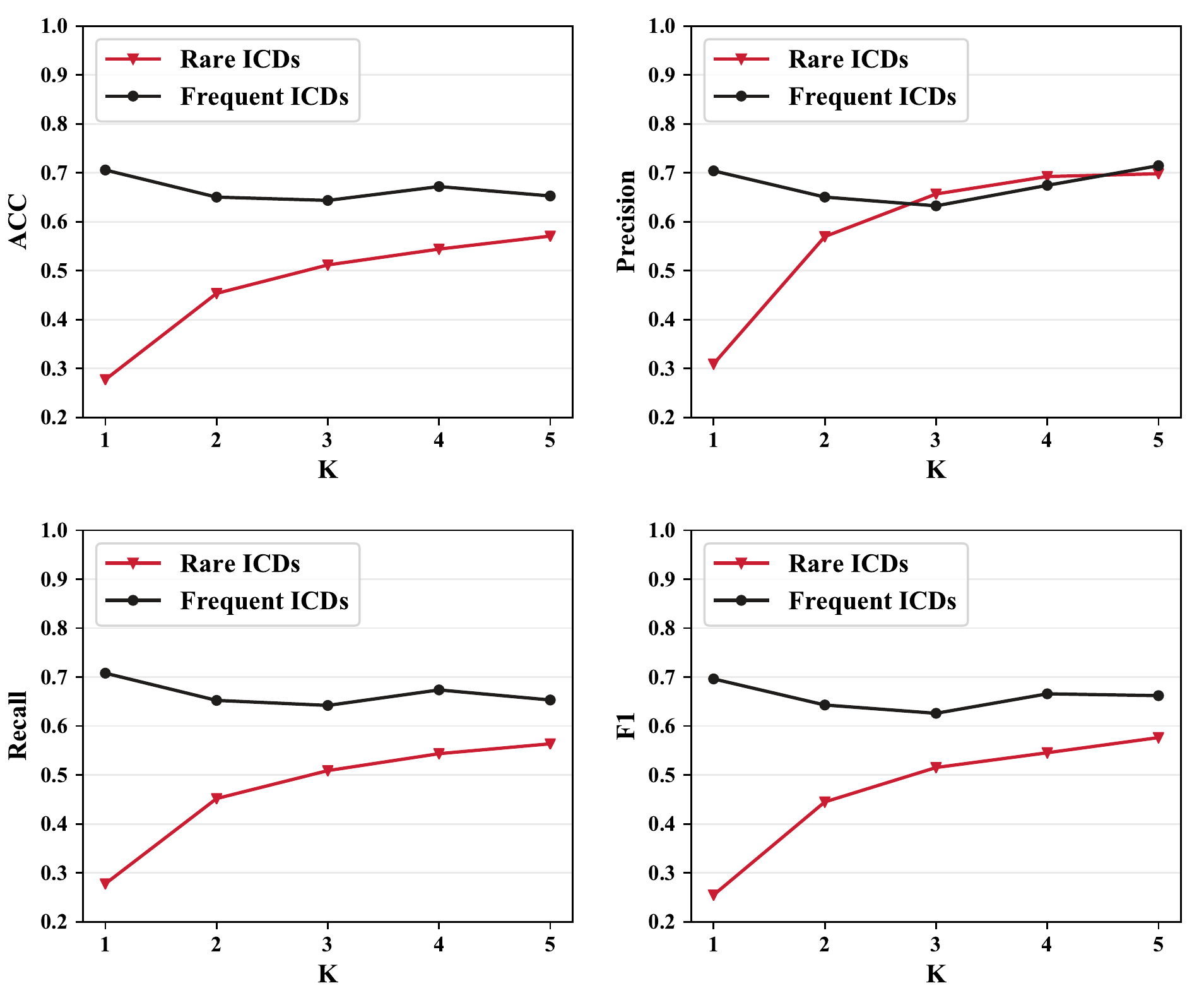}
\caption{Impact on the rare and frequent \ac{ICD} codes of few-shot size $K$.}
\label{k_size}
\end{figure}
This work studies few-shot \ac{EHR} coding, thus we conduct experiments to analyze the impact of the few-shot size $K$. 
Figure~\ref{k_size} reports the performance of \OurMethod{} on rare and frequent \ac{ICD} codes with different settings of $K$.
First, with an increase in $K$, the performances of \OurMethod{} on rare \ac{ICD} codes shows an increase. 
Specifically, when $K$ increases from 1 to 5, ACC, Precision, Recall, and F1 all show an upward trend, in which the F1 increases from 25.48\% to 57.64\%.
This indicates that a larger support set can produce better prediction results of the rare \ac{ICD} codes. 
A larger support set can provide more supervision information for query \ac{HEWE} graphs from rare \ac{ICD} codes, which is conducive to the classification of query samples~\cite{DBLP:conf/cikm/DingWLSLL20}. 

Second, when the support size $K$ increases from 1 to 2, the performance of \OurMethod{} has the greatest improvement.
For example, when $K$ increases from 1 to 4, the F1 increases by 19.01\%, 6.72\%, 3.33\%.
Further, when $K$ increases from 4 to 5, the F1 value increases by 3.10\%.
So this upward trend of model’s performance on rare \ac{ICD} codes becomes lightly gradually with the increase of $K$.
When $K$ is very small, the main factor limiting the model's performance on rare \ac{ICD} codes is the lack of supervision information.
However, with the continuous increase of $K$, the lack of supervision information is no longer its biggest limitation~\cite{DBLP:conf/eccv/SuMH20}.
Therefore, ACC, Precision, Recall, and F1 tend to increase slower as $K$ increases.

Third, the performance of \OurMethod{} on frequent \ac{ICD} codes does not increase with the increase of $K$, but has a slight downward trend.
Specifically, when $K$ increases from 1 to 3, the F1 value of \OurMethod{} gradually decreases by 5.33\% and 1.70\% respectively. 
As $K$ continues to increase, the performance of \OurMethod{} begins to fluctuate slightly.
In short, \OurMethod{} has the best performance on frequent \ac{ICD} codes when $K$ is 1.
When $K$ is 1, there is very little supervision information for the rare \ac{ICD} codes, 
which makes the model pay more attention to the classification of frequent \ac{ICD} codes.
As $K$ increases, the supervision information of rare \ac{ICD} codes prompts \OurMethod{} to learn features that can be generalized to the rare \ac{ICD} classification, so the classification performance of \OurMethod{} on frequent \ac{ICD} codes has slightly decreased.

\subsection{Impact of training \ac{ICD} classes}
\begin{figure}[t]
\centering  
\includegraphics[width=1.0\columnwidth]{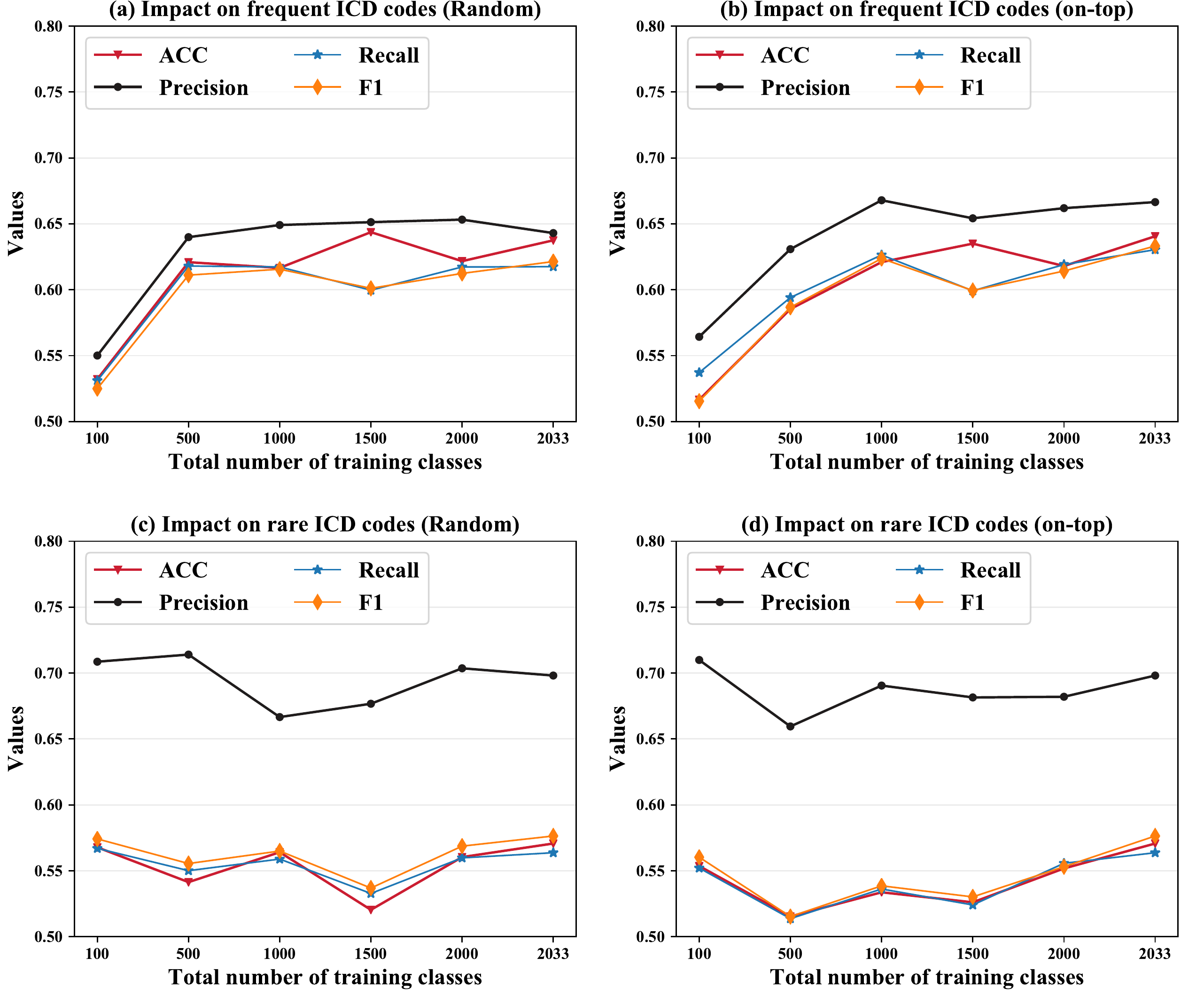}
\caption{Impact of varying number of training classes with different sampling modes.}
\label{train_size}
\end{figure}
\OurMethod{} iteratively samples C-way different \ac{ICD} classes for training, and the remaining \ac{ICD} classes are as the support set and query set. 
Suggested by ~\cite{DBLP:journals/corr/abs-2006-11325}, we conduct an analysis experiment to analyze the effect of the sampling strategies as well as the number of \ac{ICD} classes on the performance, and the results are shown in Fig~\ref{train_size}.

Fig~\ref{train_size} (a) and (b) illustrate the changing trends of \OurMethod{}'s performance on frequent \ac{ICD} codes with different sampling strategies, i.e, ``random'' and ``on-top''.
The former selects the \ac{ICD} classes for training randomly, while the latter selects in descending order by the number of samples in each class.
It can be seen that no matter which sampling strategy is used, the performances of \OurMethod{} on frequent \ac{ICD} codes rise first and then stabilize. 
\OurMethod{} fluctuates more with the ``on-top'' strategy relatively.
We believe that the main reason is that the ``on-top'' strategy is an uneven sampling strategy where \ac{ICD} codes with a large number of samples are easy to dominate, leading to unstable learning eventually.

Fig~\ref{train_size} (c) and (d) illustrate the changing trends of \OurMethod{}'s performance on rare \ac{ICD} codes with ``random'' and ``on-top'' sampling strategies.
We can observe that the overall performance of \OurMethod{} remains stable with slight fluctuations, which indicates that the increase of the training classes has little influence on the performance of \OurMethod{} in terms of the rare \ac{ICD} codes.
This means that \OurMethod{} trained with 100 \ac{ICD} codes already has sufficient learning and transfer the necessary knowledge from frequent \ac{ICD} codes to rare \ac{ICD} codes.
In addition, the Precision values of \OurMethod{} are significantly higher than Recall and ACC values for the rare \ac{ICD} codes.
This is because that the extremely rare \ac{ICD} codes are easy to be neglected, indicating that there is still room for further improvement on the extremely rare \ac{ICD} codes.

%% file: 08-conclusion.tex

\section{Conclusion}
In this work, we reformulate \acf{EHR} coding as a few-shot learning task, and propose a graph-based few-shot learning framework, i.e., \OurMethod{}, to perform \ac{ICD} classification on both frequent and rare \ac{ICD} codes.
As part of \OurMethod{}, we design two graph contrastive learning schemes, i.e., \ac{GSCL} and \ac{GSCL}, to pre-train the graph encoder in the few-shot learning framework.
\ac{GSCL} captures the intra-correlations within the graphs by comparing the structured features of sub-graphs, while \ac{GECL} captures the inter-correlations between the graphs at different clinical stages by comparing the historical graph sequence and the future graph.

A limitation of \OurMethod{} is that it only extracts the high-frequency words and entities from free-text of \acp{EHR}; it does not take medical knowledge or rules into account.
However, medical knowledge or rules may help to perform \ac{ICD} classification, such as the concurrency and hierarchy of some \ac{ICD} codes. 
As to future work, we plan to model the correlations of \ac{ICD} codes to learn better transferable features.
Then, we hope to further improve the classification performances of \OurMethod{}, especially on the rare \ac{ICD} codes.

%% file: acknowledge.tex
\section*{Acknowledgements}

This research was partially supported by the Natural Science Foundation of China (61972234, 61902219, 62072279), the National Key R\&D Program of China with grant No. 2020YFB1406704, the Key Scientific and Technological Innovation Program of Shandong Province (2019JZZY010129), the Tencent WeChat Rhino-Bird Focused Research Program (JR-WXG-2021411), the Fundamental Research Funds of Shandong University, and the Hybrid Intelligence Center, a 10-year programme funded by the Dutch Ministry of Education, Culture and Science through the Netherlands Organisation for Scientific Research, https://hybrid-intelligence-centre.nl. All content represents the opinion of the authors, which is not necessarily shared or endorsed by their respective employers and/or sponsors.

%% file: main.bbl
\begin{thebibliography}{62}
\providecommand{\natexlab}[1]{#1}
\providecommand{\url}[1]{\texttt{#1}}
\expandafter\ifx\csname urlstyle\endcsname\relax
  \providecommand{\doi}[1]{doi: #1}\else
  \providecommand{\doi}{doi: \begingroup \urlstyle{rm}\Url}\fi

\bibitem[Avati et~al.(2017)Avati, Jung, Harman, Downing, Ng, and
  Shah]{DBLP:conf/bibm/AvatiJHDNS17}
Anand Avati, Kenneth Jung, Stephanie Harman, Lance Downing, Andrew~Y. Ng, and
  Nigam~H. Shah.
\newblock Improving palliative care with deep learning.
\newblock In \emph{{IEEE} International Conference on Bioinformatics and
  Biomedicine}, pages 311--316, 2017.

\bibitem[Baumel et~al.(2018)Baumel, Nassour{-}Kassis, Cohen, Elhadad, and
  Elhadad]{DBLP:conf/aaai/BaumelNCEE18}
Tal Baumel, Jumana Nassour{-}Kassis, Raphael Cohen, Michael Elhadad, and
  No{\'{e}}mie Elhadad.
\newblock Multi-label classification of patient notes: Case study on {ICD} code
  assignment.
\newblock In \emph{The Workshops of the The Thirty-Second {AAAI} Conference on
  Artificial Intelligence}, pages 409--416, 2018.

\bibitem[Chen et~al.(2017)Chen, Sun, Shi, and Hong]{DBLP:conf/kdd/ChenSSH17}
Ting Chen, Yizhou Sun, Yue Shi, and Liangjie Hong.
\newblock On sampling strategies for neural network-based collaborative
  filtering.
\newblock In \emph{Proceedings of the 23rd {ACM} {SIGKDD} International
  Conference on Knowledge Discovery and Data Mining}, pages 767--776, 2017.

\bibitem[Chen et~al.(2020)Chen, Kornblith, Norouzi, and
  Hinton]{DBLP:conf/icml/ChenK0H20}
Ting Chen, Simon Kornblith, Mohammad Norouzi, and Geoffrey~E. Hinton.
\newblock A simple framework for contrastive learning of visual
  representations.
\newblock In \emph{Proceedings of the 37th International Conference on Machine
  Learning}, pages 1597--1607, 2020.

\bibitem[Choi et~al.(2016)Choi, Bahadori, Schuetz, Stewart, and
  Sun]{mlhc/ChoiBSSS16}
Edward Choi, Mohammad~Taha Bahadori, Andy Schuetz, Walter~F. Stewart, and
  Jimeng Sun.
\newblock Doctor {AI:} predicting clinical events via recurrent neural
  networks.
\newblock In \emph{Proceedings of the 1st Machine Learning in Health Care},
  pages 301--318, 2016.

\bibitem[Ding et~al.(2020)Ding, Wang, Li, Shu, Liu, and
  Liu]{DBLP:conf/cikm/DingWLSLL20}
Kaize Ding, Jianling Wang, Jundong Li, Kai Shu, Chenghao Liu, and Huan Liu.
\newblock Graph prototypical networks for few-shot learning on attributed
  networks.
\newblock In \emph{The 29th {ACM} International Conference on Information and
  Knowledge Management}, pages 295--304, 2020.

\bibitem[Finn et~al.(2017)Finn, Abbeel, and Levine]{DBLP:conf/icml/FinnAL17}
Chelsea Finn, Pieter Abbeel, and Sergey Levine.
\newblock Model-agnostic meta-learning for fast adaptation of deep networks.
\newblock In \emph{Proceedings of the 34th International Conference on Machine
  Learning}, pages 1126--1135, 2017.

\bibitem[Fukui et~al.(2016)Fukui, Park, Yang, Rohrbach, Darrell, and
  Rohrbach]{fukui2016multimodal}
Akira Fukui, Dong~Huk Park, Daylen Yang, Anna Rohrbach, Trevor Darrell, and
  Marcus Rohrbach.
\newblock Multimodal compact bilinear pooling for visual question answering and
  visual grounding.
\newblock In \emph{Proceedings of the 2016 Conference on Empirical Methods in
  Natural Language Processing}, pages 457--468, 2016.

\bibitem[Gao et~al.(2021)Gao, Yao, and Chen]{DBLP:journals/corr/abs-2104-08821}
Tianyu Gao, Xingcheng Yao, and Danqi Chen.
\newblock Simcse: Simple contrastive learning of sentence embeddings.
\newblock \emph{CoRR}, abs/2104.08821, 2021.

\bibitem[Geng et~al.(2020)Geng, Li, Li, Sun, and Zhu]{DBLP:conf/acl/GengLLSZ20}
Ruiying Geng, Binhua Li, Yongbin Li, Jian Sun, and Xiaodan Zhu.
\newblock Dynamic memory induction networks for few-shot text classification.
\newblock In \emph{Proceedings of the 58th Annual Meeting of the Association
  for Computational Linguistics}, pages 1087--1094, 2020.

\bibitem[Goutte and Gaussier(2005)]{goutte2005probabilistic}
Cyril Goutte and Eric Gaussier.
\newblock A probabilistic interpretation of precision, recall and f-score, with
  implication for evaluation.
\newblock In \emph{European Conference on Information Retrieval}, pages
  345--359, 2005.

\bibitem[He et~al.(2020)He, Fan, Wu, Xie, and
  Girshick]{DBLP:conf/cvpr/He0WXG20}
Kaiming He, Haoqi Fan, Yuxin Wu, Saining Xie, and Ross~B. Girshick.
\newblock Momentum contrast for unsupervised visual representation learning.
\newblock In \emph{{IEEE/CVF} Conference on Computer Vision and Pattern
  Recognition}, pages 9726--9735, 2020.

\bibitem[Huang and Ling(2005)]{huang2005using}
Jin Huang and Charles~X Ling.
\newblock Using auc and accuracy in evaluating learning algorithms.
\newblock \emph{IEEE Transactions on knowledge and Data Engineering},
  17\penalty0 (3):\penalty0 299--310, 2005.

\bibitem[Huang et~al.(2019)Huang, Osorio, and Sy]{DBLP:journals/cmpb/HuangOS19}
Jinmiao Huang, Cesar Osorio, and Luke~Wicent Sy.
\newblock An empirical evaluation of deep learning for {ICD-9} code assignment
  using {MIMIC-III} clinical notes.
\newblock \emph{Computer Methods and Programs in Biomedicine}, 177:\penalty0
  141--153, 2019.

\bibitem[Jeh and Widom(2003)]{DBLP:conf/www/JehW03}
Glen Jeh and Jennifer Widom.
\newblock Scaling personalized web search.
\newblock In \emph{Proceedings of the Twelfth International World Wide Web
  Conference}, pages 271--279, 2003.

\bibitem[Jiao et~al.(2020)Jiao, Xiong, Zhang, Zhang, Zhang, and
  Zhu]{jiao2020sub}
Yizhu Jiao, Yun Xiong, Jiawei Zhang, Yao Zhang, Tianqi Zhang, and Yangyong Zhu.
\newblock Sub-graph contrast for scalable self-supervised graph representation
  learning.
\newblock \emph{arXiv preprint arXiv:2009.10273}, 2020.

\bibitem[Johnson et~al.(2016)Johnson, Pollard, Shen, Li-Wei, Feng, Ghassemi,
  Moody, Szolovits, Celi, and Mark]{johnson2016mimic}
Alistair~EW Johnson, Tom~J Pollard, Lu~Shen, H~Lehman Li-Wei, Mengling Feng,
  Mohammad Ghassemi, Benjamin Moody, Peter Szolovits, Leo~Anthony Celi, and
  Roger~G Mark.
\newblock Mimic-iii, a freely accessible critical care database.
\newblock \emph{Scientific data}, 3\penalty0 (1):\penalty0 1--9, 2016.

\bibitem[Kim(2014)]{emnlp/Kim14}
Yoon Kim.
\newblock Convolutional neural networks for sentence classification.
\newblock In \emph{Proceedings of Conference on Empirical Methods in Natural
  Language Processing}, pages 1746--1751, 2014.

\bibitem[Kingma and Ba(2015)]{DBLP:journals/corr/KingmaB14}
Diederik~P. Kingma and Jimmy Ba.
\newblock Adam: {A} method for stochastic optimization.
\newblock In \emph{3rd International Conference on Learning Representations},
  2015.

\bibitem[Kipf and Welling(2017)]{DBLP:conf/iclr/KipfW17}
Thomas~N. Kipf and Max Welling.
\newblock Semi-supervised classification with graph convolutional networks.
\newblock In \emph{5th International Conference on Learning Representations},
  2017.

\bibitem[Li et~al.(2020)Li, Zhang, Wei, and Zhang]{DBLP:conf/igarss/LiZWZ20}
Yu~Li, Lei Zhang, Wei Wei, and Yanning Zhang.
\newblock Deep self-supervised learning for few-shot hyperspectral image
  classification.
\newblock In \emph{{IEEE} International Geoscience and Remote Sensing
  Symposium}, pages 501--504, 2020.

\bibitem[Li et~al.(2017)Li, Zhou, Chen, and Li]{DBLP:journals/corr/LiZCL17}
Zhenguo Li, Fengwei Zhou, Fei Chen, and Hang Li.
\newblock Meta-sgd: Learning to learn quickly for few shot learning.
\newblock \emph{CoRR}, abs/1707.09835, 2017.

\bibitem[Madotto et~al.(2019)Madotto, Lin, Wu, and
  Fung]{DBLP:conf/acl/MadottoLWF19}
Andrea Madotto, Zhaojiang Lin, Chien{-}Sheng Wu, and Pascale Fung.
\newblock Personalizing dialogue agents via meta-learning.
\newblock In \emph{Proceedings of the 57th Conference of the Association for
  Computational Linguistics}, pages 5454--5459, 2019.

\bibitem[Medina et~al.(2020)Medina, Devos, and
  Grossglauser]{DBLP:journals/corr/abs-2006-11325}
Carlos Medina, Arnout Devos, and Matthias Grossglauser.
\newblock Self-supervised prototypical transfer learning for few-shot
  classification.
\newblock \emph{CoRR}, abs/2006.11325, 2020.

\bibitem[Mikolov et~al.(2013)Mikolov, Sutskever, Chen, Corrado, and
  Dean]{NIPS2013_9aa42b31}
Tomas Mikolov, Ilya Sutskever, Kai Chen, Greg~S Corrado, and Jeff Dean.
\newblock Distributed representations of words and phrases and their
  compositionality.
\newblock In \emph{Advances in Neural Information Processing Systems},
  volume~26, pages 3111--3119, 2013.

\bibitem[Mishra et~al.(2018)Mishra, Rohaninejad, Chen, and
  Abbeel]{DBLP:conf/iclr/MishraR0A18}
Nikhil Mishra, Mostafa Rohaninejad, Xi~Chen, and Pieter Abbeel.
\newblock A simple neural attentive meta-learner.
\newblock In \emph{6th International Conference on Learning Representations},
  2018.

\bibitem[Mullenbach et~al.(2018{\natexlab{a}})Mullenbach, Wiegreffe, Duke, Sun,
  and Eisenstein]{DBLP:conf/naacl/MullenbachWDSE18}
James Mullenbach, Sarah Wiegreffe, Jon Duke, Jimeng Sun, and Jacob Eisenstein.
\newblock Explainable prediction of medical codes from clinical text.
\newblock In \emph{Proceedings of the 2018 Conference of the North American
  Chapter of the Association for Computational Linguistics: Human Language
  Technologies}, pages 1101--1111, 2018{\natexlab{a}}.

\bibitem[Mullenbach et~al.(2018{\natexlab{b}})Mullenbach, Wiegreffe, Duke, Sun,
  and Eisenstein]{naacl/MullenbachWDSE18}
James Mullenbach, Sarah Wiegreffe, Jon Duke, Jimeng Sun, and Jacob Eisenstein.
\newblock Explainable prediction of medical codes from clinical text.
\newblock In \emph{Proceedings of the 2018 Conference of the North American
  Chapter of the Association for Computational Linguistics: Human Language
  Technologies}, pages 1101--1111, 2018{\natexlab{b}}.

\bibitem[Peng et~al.(2020)Peng, Dong, Luo, Wu, and
  Zheng]{DBLP:journals/corr/abs-2003-01604}
Zhen Peng, Yixiang Dong, Minnan Luo, Xiao{-}Ming Wu, and Qinghua Zheng.
\newblock Self-supervised graph representation learning via global context
  prediction.
\newblock \emph{CoRR}, abs/2003.01604, 2020.

\bibitem[Prakash et~al.(2017)Prakash, Zhao, Hasan, Datla, Lee, Qadir, Liu, and
  Farri]{aaai/PrakashZHDLQLF17}
Aaditya Prakash, Siyuan Zhao, Sadid~A. Hasan, Vivek~V. Datla, Kathy Lee,
  Ashequl Qadir, Joey Liu, and Oladimeji Farri.
\newblock Condensed memory networks for clinical diagnostic inferencing.
\newblock In \emph{Proceedings of the Thirty-First Conference on Artificial
  Intelligence}, pages 3274--3280, 2017.

\bibitem[Qin et~al.(2020)Qin, Li, Shi, and
  Gao]{DBLP:journals/corr/abs-2004-05805}
Tiexin Qin, Wenbin Li, Yinghuan Shi, and Yang Gao.
\newblock Unsupervised few-shot learning via distribution shift-based
  augmentation.
\newblock \emph{CoRR}, abs/2004.05805, 2020.

\bibitem[Qiu et~al.(2020)Qiu, Chen, Dong, Zhang, Yang, Ding, Wang, and
  Tang]{DBLP:conf/kdd/QiuCDZYDWT20}
Jiezhong Qiu, Qibin Chen, Yuxiao Dong, Jing Zhang, Hongxia Yang, Ming Ding,
  Kuansan Wang, and Jie Tang.
\newblock {GCC:} graph contrastive coding for graph neural network
  pre-training.
\newblock In \emph{The 26th {ACM} {SIGKDD} Conference on Knowledge Discovery
  and Data Mining}, pages 1150--1160, 2020.

\bibitem[Ravi and Larochelle(2017)]{iclr/RaviL17}
Sachin Ravi and Hugo Larochelle.
\newblock Optimization as a model for few-shot learning.
\newblock In \emph{5th International Conference on Learning Representations},
  2017.

\bibitem[Roy et~al.(2020)Roy, Xu, Wang, Kitani, Salakhutdinov, and
  Hebert]{DBLP:journals/corr/abs-2008-09892}
Vivek Roy, Yan Xu, Yu{-}Xiong Wang, Kris Kitani, Ruslan Salakhutdinov, and
  Martial Hebert.
\newblock Few-shot learning with intra-class knowledge transfer.
\newblock \emph{CoRR}, abs/2008.09892, 2020.

\bibitem[Sbai et~al.(2020)Sbai, Couprie, and Aubry]{sbai2020impact}
Othman Sbai, Camille Couprie, and Mathieu Aubry.
\newblock Impact of base dataset design on few-shot image classification.
\newblock In \emph{Computer Vision - 16th European Conference}, pages 597--613,
  2020.

\bibitem[Shi et~al.(2017)Shi, Xie, Hu, Zhang, and
  Xing]{DBLP:journals/corr/abs-1711-04075}
Haoran Shi, Pengtao Xie, Zhiting Hu, Ming Zhang, and Eric~P. Xing.
\newblock Towards automated {ICD} coding using deep learning.
\newblock \emph{CoRR}, abs/1711.04075, 2017.

\bibitem[Snell et~al.(2017{\natexlab{a}})Snell, Swersky, and
  Zemel]{snell2017prototypical}
Jake Snell, Kevin Swersky, and Richard Zemel.
\newblock Prototypical networks for few-shot learning.
\newblock In \emph{Advances in neural information processing systems}, pages
  4077--4087, 2017{\natexlab{a}}.

\bibitem[Snell et~al.(2017{\natexlab{b}})Snell, Swersky, and
  Zemel]{DBLP:conf/nips/SnellSZ17}
Jake Snell, Kevin Swersky, and Richard~S. Zemel.
\newblock Prototypical networks for few-shot learning.
\newblock In \emph{Annual Conference on Neural Information Processing Systems},
  pages 4077--4087, 2017{\natexlab{b}}.

\bibitem[Sohn(2016)]{DBLP:conf/nips/Sohn16}
Kihyuk Sohn.
\newblock Improved deep metric learning with multi-class n-pair loss objective.
\newblock In \emph{Annual Conference on Neural Information Processing Systems},
  pages 1849--1857, 2016.

\bibitem[Su et~al.(2019)Su, Maji, and
  Hariharan]{DBLP:journals/corr/abs-1910-03560}
Jong{-}Chyi Su, Subhransu Maji, and Bharath Hariharan.
\newblock When does self-supervision improve few-shot learning?
\newblock \emph{CoRR}, abs/1910.03560, 2019.

\bibitem[Su et~al.(2020)Su, Maji, and Hariharan]{DBLP:conf/eccv/SuMH20}
Jong{-}Chyi Su, Subhransu Maji, and Bharath Hariharan.
\newblock When does self-supervision improve few-shot learning?
\newblock In \emph{Computer Vision - 16th European Conference}, pages 645--666,
  2020.

\bibitem[Sung et~al.(2018{\natexlab{a}})Sung, Yang, Zhang, Xiang, Torr, and
  Hospedales]{DBLP:conf/cvpr/SungYZXTH18}
Flood Sung, Yongxin Yang, Li~Zhang, Tao Xiang, Philip H.~S. Torr, and
  Timothy~M. Hospedales.
\newblock Learning to compare: Relation network for few-shot learning.
\newblock In \emph{{IEEE} Conference on Computer Vision and Pattern
  Recognition}, pages 1199--1208, 2018{\natexlab{a}}.

\bibitem[Sung et~al.(2018{\natexlab{b}})Sung, Yang, Zhang, Xiang, Torr, and
  Hospedales]{sung2018learning}
Flood Sung, Yongxin Yang, Li~Zhang, Tao Xiang, Philip~HS Torr, and Timothy~M
  Hospedales.
\newblock Learning to compare: Relation network for few-shot learning.
\newblock In \emph{Proceedings of the IEEE Conference on Computer Vision and
  Pattern Recognition}, pages 1199--1208, 2018{\natexlab{b}}.

\bibitem[Teng et~al.(2020)Teng, Yang, Chen, Huang, and Xu]{frontiers/Teng2020}
Fei Teng, Wei Yang, Li~Chen, LuFei Huang, and Qiang Xu.
\newblock Explainable prediction of medical codes with knowledge graphs.
\newblock \emph{Frontiers in Bioengineering and Biotechnology}, 8:\penalty0
  867, 2020.

\bibitem[Tian et~al.(2019)Tian, Krishnan, and
  Isola]{DBLP:journals/corr/abs-1906-05849}
Yonglong Tian, Dilip Krishnan, and Phillip Isola.
\newblock Contrastive multiview coding.
\newblock \emph{CoRR}, abs/1906.05849, 2019.

\bibitem[van~den Oord et~al.(2018)van~den Oord, Li, and
  Vinyals]{DBLP:journals/corr/abs-1807-03748}
A{\"{a}}ron van~den Oord, Yazhe Li, and Oriol Vinyals.
\newblock Representation learning with contrastive predictive coding.
\newblock \emph{CoRR}, abs/1807.03748, 2018.

\bibitem[van~der Maaten and Hinton(2008)]{van2008visualizing}
Laurens van~der Maaten and Geoffrey Hinton.
\newblock Visualizing data using t-sne.
\newblock \emph{Journal of Machine Learning Research}, 9\penalty0
  (86):\penalty0 2579--2605, 2008.

\bibitem[Velickovic et~al.(2019)Velickovic, Fedus, Hamilton, Li{\`{o}}, Bengio,
  and Hjelm]{DBLP:conf/iclr/VelickovicFHLBH19}
Petar Velickovic, William Fedus, William~L. Hamilton, Pietro Li{\`{o}}, Yoshua
  Bengio, and R.~Devon Hjelm.
\newblock Deep graph infomax.
\newblock In \emph{7th International Conference on Learning Representations},
  2019.

\bibitem[Vinyals et~al.(2016)Vinyals, Blundell, Lillicrap, Kavukcuoglu, and
  Wierstra]{DBLP:conf/nips/VinyalsBLKW16}
Oriol Vinyals, Charles Blundell, Tim Lillicrap, Koray Kavukcuoglu, and Daan
  Wierstra.
\newblock Matching networks for one shot learning.
\newblock In \emph{Annual Conference on Neural Information Processing Systems},
  pages 3630--3638, 2016.

\bibitem[Wang et~al.(2016)Wang, Chang, Li, Long, Yao, and
  Sheng]{DBLP:journals/tkde/WangCLLYS16}
Sen Wang, Xiaojun Chang, Xue Li, Guodong Long, Lina Yao, and Quan~Z. Sheng.
\newblock Diagnosis code assignment using sparsity-based disease correlation
  embedding.
\newblock \emph{IEEE Transactions on Knowledge and Data Engineering},
  28\penalty0 (12):\penalty0 3191--3202, 2016.

\bibitem[Wang et~al.(2019)Wang, Ren, Chen, Ren, Ma, and
  de~Rijke]{10.1145/3357384.3357965}
Shanshan Wang, Pengjie Ren, Zhumin Chen, Zhaochun Ren, Jun Ma, and Maarten
  de~Rijke.
\newblock Order-free medicine combination prediction with graph convolutional
  reinforcement learning.
\newblock In \emph{Proceedings of the 28th ACM International Conference on
  Information and Knowledge Management}, page 1623–1632, 2019.

\bibitem[Wang et~al.(2020)Wang, Ren, Chen, Ren, Nie, Ma, and
  de~Rijke]{wang-2020-coding}
Shanshan Wang, Pengjie Ren, Zhumin Chen, Zhaochun Ren, Jian-Yun Nie, Jun Ma,
  and Maarten de~Rijke.
\newblock Coding electronic health records with adversarial reinforcement path
  generation.
\newblock In \emph{43rd international ACM SIGIR conference on Research and
  Development in Information Retrieval}, pages 801--810, 2020.

\bibitem[Wu et~al.(2018)Wu, Xiong, Yu, and Lin]{DBLP:conf/cvpr/WuXYL18}
Zhirong Wu, Yuanjun Xiong, Stella~X. Yu, and Dahua Lin.
\newblock Unsupervised feature learning via non-parametric instance
  discrimination.
\newblock In \emph{{IEEE} Conference on Computer Vision and Pattern
  Recognition}, pages 3733--3742, 2018.

\bibitem[Xie et~al.(2019)Xie, Xiong, Yu, and Zhu]{cikm/Xie18}
Xiancheng Xie, Yun Xiong, Philip~S. Yu, and Yangyong Zhu.
\newblock Ehr coding with multi-scale feature attention and structured
  knowledge graph propagation.
\newblock In \emph{Proceedings of the 28th ACM International Conference on
  Information and Knowledge Management}, pages 649--658, 2019.

\bibitem[Xu et~al.(2019)Xu, Lam, Pang, Gao, Band, Mathur, Papay, Khanna,
  Cywinski, Maheshwari, Xie, and Xing]{DBLP:conf/mlhc/XuLPGBMPKCMXX19}
Keyang Xu, Mike Lam, Jingzhi Pang, Xin Gao, Charlotte Band, Piyush Mathur,
  Frank Papay, Ashish~K. Khanna, Jacek~B. Cywinski, Kamal Maheshwari, Pengtao
  Xie, and Eric~P. Xing.
\newblock Multimodal machine learning for automated {ICD} coding.
\newblock In \emph{Proceedings of the Machine Learning for Healthcare
  Conference}, pages 197--215, 2019.

\bibitem[Yang and Xu(2020)]{NEURIPS2020_e025b627}
Yuzhe Yang and Zhi Xu.
\newblock Rethinking the value of labels for improving class-imbalanced
  learning.
\newblock In \emph{Advances in Neural Information Processing Systems},
  volume~33, pages 19290--19301, 2020.

\bibitem[Ye et~al.(2019)Ye, Zhang, Yuen, and Chang]{DBLP:conf/cvpr/YeZYC19}
Mang Ye, Xu~Zhang, Pong~C. Yuen, and Shih{-}Fu Chang.
\newblock Unsupervised embedding learning via invariant and spreading instance
  feature.
\newblock In \emph{{IEEE} Conference on Computer Vision and Pattern
  Recognition}, pages 6210--6219, 2019.

\bibitem[Yu et~al.(2019)Yu, Zhao, and Wang]{DBLP:conf/aipr2/YuZW19}
Qing Yu, Hui Zhao, and Zuohua Wang.
\newblock Attention-based bidirectional gated recurrent unit neural networks
  for sentiment analysis.
\newblock In \emph{Proceedings of the 2nd International Conference on
  Artificial Intelligence and Pattern Recognition}, pages 116--119, 2019.

\bibitem[Zeng and Xie(2020)]{DBLP:journals/corr/abs-2009-05923}
Jiaqi Zeng and Pengtao Xie.
\newblock Contrastive self-supervised learning for graph classification.
\newblock \emph{CoRR}, abs/2009.05923, 2020.

\bibitem[Zhang et~al.(2017)Zhang, He, Zhao, and
  Li]{DBLP:conf/bionlp/ZhangHZL17}
Danchen Zhang, Daqing He, Sanqiang Zhao, and Lei Li.
\newblock Enhancing automatic {ICD-9-CM} code assignment for medical texts with
  pubmed.
\newblock In \emph{Workshop on Biomedical Natural Language Processing}, pages
  263--271, 2017.

\bibitem[Zhang et~al.(2020)Zhang, Zhang, Xia, and
  Sun]{DBLP:journals/corr/abs-2001-05140}
Jiawei Zhang, Haopeng Zhang, Congying Xia, and Li~Sun.
\newblock Graph-bert: Only attention is needed for learning graph
  representations.
\newblock \emph{CoRR}, abs/2001.05140, 2020.

\bibitem[Zhou et~al.(2016)Zhou, Shi, Tian, Qi, Li, Hao, and
  Xu]{acl/ZhouSTQLHX16}
Peng Zhou, Wei Shi, Jun Tian, Zhenyu Qi, Bingchen Li, Hongwei Hao, and Bo~Xu.
\newblock Attention-based bidirectional long short-term memory networks for
  relation classification.
\newblock In \emph{Proceedings of the 54th Annual Meeting of the Association
  for Computational Linguistics}, pages 207--212, 2016.

\end{thebibliography}
